\documentclass[sigconf]{aamas}

\usepackage{balance} 
\usepackage{algorithm}
\usepackage[noend]{algpseudocode}
\usepackage{multirow}

\doi{RJIB1203}

\makeatletter
\gdef\@copyrightpermission{
  \begin{minipage}{0.2\columnwidth}
   \href{https://creativecommons.org/licenses/by/4.0/}{\includegraphics[width=0.90\textwidth]{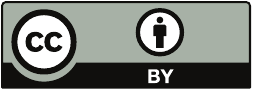}}
  \end{minipage}\hfill
  \begin{minipage}{0.8\columnwidth}
   \href{https://creativecommons.org/licenses/by/4.0/}{This work is licensed under a Creative Commons Attribution International 4.0 License.}
  \end{minipage}
  \vspace{5pt}
}
\makeatother

\setcopyright{ifaamas}
\acmConference[AAMAS '26]{Proc.\@ of the 25th International Conference
on Autonomous Agents and Multiagent Systems (AAMAS 2026)}{May 25 -- 29, 2026}
{Paphos, Cyprus}{C.~Amato, L.~Dennis, V.~Mascardi, J.~Thangarajah (eds.)}
\copyrightyear{2026}
\acmYear{2026}
\acmDOI{}
\acmPrice{}
\acmISBN{}

\acmSubmissionID{990}

\title[$Q$ Inverse Constrained RL]{Learning to maintain safety through expert demonstrations in settings with unknown constraints: A  Q-learning perspective.}

\author{George Papadopoulos}
\affiliation{
  \institution{University of Piraeus}
  \city{Piraeus}
  \country{Greece}}
\email{georgepap@unipi.gr}

\author{George A. Vouros}
\affiliation{
  \institution{University of Piraeus}
  \city{Piraeus}
  \country{Greece}}
\email{georgev@unipi.gr}

\begin{abstract}
Given a set of trajectories demonstrating the execution of a task safely in a constrained MDP with observable rewards but with unknown constraints and non-observable costs, we aim to find a policy that maximizes the likelihood of demonstrated trajectories trading the balance between being conservative and increasing significantly the likelihood of high-rewarding trajectories but with potentially unsafe steps. Having these objectives, we aim towards learning a policy that maximizes the probability of the most $promising$ trajectories with respect to the demonstrations. In so doing, we formulate the ``promise" of individual state-action pairs in terms of  $Q$ values, which depend on task-specific rewards as well as on the assessment of states' safety, mixing expectations in terms of rewards and safety. This entails a safe Q-learning perspective of the inverse learning problem under constraints: The devised Safe $Q$ Inverse Constrained Reinforcement Learning (SafeQIL) algorithm is compared to state-of-the art inverse constraint reinforcement learning algorithms to a set of challenging benchmark tasks, showing its merits. 
\end{abstract}


\newcommand{\BibTeX}{\rm B\kern-.05em{\sc i\kern-.025em b}\kern-.08em\TeX}

\begin{document}


\pagestyle{fancy}
\fancyhead{}


\maketitle 


\section{Introduction}

In this paper, we address the inverse problem of learning safe policies given expert trajectories following underlying constraints and demonstrating safe execution of tasks under observable rewards and non-observable costs. We term this as an inverse learning problem with respect to constraints, given that the set of constraints are unknown, and although we do not aim to approximate the set of constraints or cost functions that determine the set of expert demonstrations safe, we aim to assess the safety of states and learn policies that support agents to act safely not only by  following the demonstrated trajectories, but also in states not in the support of those demonstrated. This  perspective distinguishes this work from inverse constrained reinforcement learning approaches that aim to recover the minimal constraint sets or the least constraining constraints explaining the demonstrated behavior \cite{pmlr-v139-malik21a, liu2023benchmarking}.

Considering possible tradeoffs between rewards and constraints in executing a task, in settings with non-observable costs, agents may either learn very conservative policies that avoid states and actions not included in the demonstrated trajectories, or policies that increase significantly the likelihood of trajectories that cross states with high uncertainty about their safety but with high expected rewards. These will be the cases when, for instance, assessing the feasibility of agents' behavior at a trajectory level in a conservative manner, or when the values of state-action pairs not in the demonstrated trajectories are much larger than those in the demonstrated trajectories in terms of the expected reward. This means that in the first case the task execution will be punished as a whole and the agents shall not have ability to recover safety at any trajectory step, and in the later case the policy will prefer to cross states and perform high-rewarding actions in state-action space areas with high uncertainty about safety. 

To address these phenomena, we express the likelihood of trajectories in terms of Q-values of individual state-action pairs, mixing expectations in terms of rewards and safety. 

This approach aims towards policies that indicate the state-action pairs demonstrated to be the most likely ones, without increasing much the likelihood of trajectories that cross states that are not in the support of demonstrated state-action pairs (i.e., being probably unsafe), and without being conservative. The basic idea is that Q values mixing expected rewards and safety should indicate actions as highly promising when leading to subsequent states assessed to be safe. This supports agents to learn how to recover safety, targeting to safe states, even when they are in states of high uncertainty about their safety. 

Although costs for violating constraints, or constraints-related rewards, are not revealed explicitly or approximated, to assess states' safety we use a discriminator function that estimates the probability of a state to be included in the distribution of states demonstrated. This brings this approach closer to other inverse constrained reinforcement learning approaches \cite{pmlr-v139-malik21a, liu2023benchmarking}, but with a different objective regarding the likelihood of trajectories and significant differences in performance.

Finally, it must be noted that mixing rewards and safety in Q values is different from reward-shaping approaches, where cost functions are assumed to be known: In contrast to that, we evaluate actions performed in states on their expectation to maintain safety, and distinctly to that, in terms of their expected rewards. 

The contributions that this article makes are as follows:

\textbf{(a)} It formulates the problem of learning a policy with respect to expert (safe) demonstrated trajectories as an inverse constrained reinforcement learning problem, whose objective function is rigorously specified in terms of Q-values of trajectory steps incorporating assessments on the safety of states, mixing expectations in terms of rewards and safety.

\textbf{(b)} It proposes the safe  $Q$ Inverse Constrained Reinforcement Learning (SafeQIL) algorithm\footnote{Our implementation, the generated dataset, and the supplementary material are publicly available in: \url{https://github.com/AILabDsUnipi/SafeQIL} .}.

\textbf{(c)} It presents evaluation results for SafeQIL in settings with constraints of increasing complexity. These are compared to results from state of the art imitation and inverse constrained reinforcement learning algorithms.


\section{Preliminaries and Motivation}

A Markov Decision Process (MDP) is a tuple $(S, A, R, P, \gamma)$, where $S$ is the set of states, $A$ is the set of actions available to an agent, $r: S \times A \rightarrow \mathbb{R}$ is the reward function,  $P: S \times A \times S \rightarrow [0,1]$ is the transition probability function $P(s_{t+1}|a_t,s_t)$ to a new state $s_{t+1}$ after the execution of action $a_t$ in state $s_t$, and $\gamma \in [0,1]$ is the discount factor. 
In such a setting, an agent aims to learn the optimal policy $\pi^*: S \rightarrow \mathcal{P}(A)$ from states to probability distributions over actions, so as to maximize the performance measure 
\begin{align}
    J(\pi)=\mathbb{E}_{\tau \sim \pi} [\sum_{t=0}^\infty \gamma^tr(s_t, a_t)] + \alpha\mathcal{H}(\pi)
\end{align} 

\noindent where $\tau$ denotes any trajectory generated using the policy $\pi^*$, with $s_0 \sim \mu \text{ (the distribution of initial states), } a_t \sim \pi(\cdot|s_t), s_{t+1} \sim P(\cdot|s_t, a_t)$, and $\mathcal{H}(\pi)$ represents the policy entropy weighted by $\alpha$.

A constrained Markov decision process (CMDP) is an MDP with constraints that restrict the set of feasible policies for the MDP. Thus, the MDP is augmented with a set of constraint functions $\mathcal{C}=\{C_i, i=1..., m\}$, and corresponding limits $d_i, i=1 ..., m$.  Each $C_i:S \times A  \rightarrow \mathbb{R}$ maps the execution of actions at states to costs.  In such a setting, the set of feasible policies with respect to constraints is 
$\Pi_C=\{\pi \in \Pi | \forall i, J_{C_i}(\pi)\leq d_i\}$ \footnote{Here, $d_i, i=1,...,m$ denote trajectory-level cost limits, although in alternative formulations they may denote stepwise limits. Alternatively, we may define trajectory level constraints of the form $C_i(\tau)$, where $\tau$ is a trajectory generated by a policy $\pi$. Then $J_{C_i}(\pi)$ is of the form $\mathbb{E}_{\tau \sim \pi} [C_i(\tau)]$}
\noindent where:  
\begin{equation*}
    J_{C_i}(\pi)=\mathbb{E}_{\tau \sim \pi} [\sum_{t=0}^\infty \gamma^t C_i(s_t, a_t)]
\end{equation*}
 denotes the $i$-th constraint-related discounted cost of policy $\pi$.
 In such a setting with known or predefined constraints, the constrained reinforcement learning objective is to learn the performance $J(\cdot)$ maximizing a feasible policy. Formally,
\begin{equation*}
    \pi^*=argmax_{\pi \in \Pi_C} J(\pi).   
\end{equation*}

However, as it is well known, in many settings constraints can not be formulated or even expressed by human experts, being inherent in their expertise. Therefore, such information cannot be revealed to an agent in an explicit way, but can only be inferred through demonstrations. In such settings of unknown or implicit constraints, the inverse constrained reinforcement learning (ICRL) problem aims at fitting constraint functions $\tilde{C_i}$, given a set of sample  trajectories $\mathcal{D}_E$ generated by an expert policy $\pi_E$, solving the optimization problem: 
\[max_{\tilde{C_i}}(J_{\pi \in \Pi_{\tilde{C_i}}}(\pi)-J_{\tilde{C_i}}(\pi_E)).\] 

Such a process, as shown in \cite{liu2025comprehensivesurveyinverseconstrained}, iterates through constraint update (the inverse step) and policy update steps, given the set of expert trajectories $\mathcal{D}_E$.
ICRL aims to model the  expert policy $\pi_E$ 
\begin{equation*}
    \tilde{\pi_E}=argmax_{\pi \in \Pi_{\tilde{C}}} J(\pi)    
\end{equation*}
using a constrained reinforcement learning method, where $\tilde{C}$ is the set of approximated constraint functions $\tilde{C_i}$.

Actually, given that the demonstrated trajectories may be explained by different sets of constraints, ICRL looks for the minimal set of constraint functions that maximizes the likelihood of demonstrations. Thus, it aims to assign a low cost to the trajectories generated by the expert policy, which must be feasible (i.e., within cost limits),  and high cost to trajectories generated by any other feasible policy. Finding this set of constraints among the combinations of potential constraints is an intractable problem where greedy solutions can be used \cite{scobee2020maximumlikelihoodconstraintinference}. Instead of doing so, as proposed in \cite{pmlr-v139-malik21a}, we may define a (learnable) function $\phi: S \times A \rightarrow [0,1]$ that indicates the probability of an action performed in a state to be safe (i.e., within the distribution of expert demonstrations). This function may be used for calculating the probability of satisfying trajectory-level safety constraints $C(\tau)=1-\prod_{(s,a)\in \tau}\phi(s,a)=1-\mathbb{I}^\phi(\tau)$.

In doing so, this approach, although termed as an inverse approach, does not aim to reveal the set of constraints explaining the demonstrated trajectories, but to find the policy that increases in a direct manner the likelihood of demonstrated trajectories. This results into  training this discrimination function $\phi_\omega$, parameterized by $\omega$, with the  following objective:
\[max_\omega log p(\mathcal{D}_E|\phi)= max_\omega  log\big{[}\frac{1}{Z_\phi^{|\mathcal{D}_E|}}\prod_{i=1}^{|\mathcal{D}_E|}exp[r(\tau^i)]\mathbb{I}^\phi(\tau^i) \big{]}\] 
\[\text{ with } Z_\phi=\int \mathbb{I}^\phi(\tau)exp(r(\tau))\,d\tau, \text{ and } \mathbb{I}^\phi(\tau^i)=\prod_{t=1}^T\phi_\omega(s_t^i, a_t^i)\]

However, formulating the objective in this way implies the following:
\begin{itemize}
    \item The assessment of the feasibility of any trajectory by means of $\mathbb{I}^\phi(\tau)$ is very strict, given that in case a single step (state-action pair) is not in the distribution of those demonstrated, this can be considered to be unsafe, and this may result into $C(\tau) \text{ close to} 1$. This may result to a very conservative agent behavior. 
    \item Highly rewarding trajectories, beyond those demonstrated, with state-action pairs to which $\phi_\omega$ is approximately 0.5 may lead the agent to unsafe behavior, considering the trajectory as highly promising.
    \item Indeed, from a stepwise perspective, there may exist individual state-action pairs to which the agent is uncertain about their safety, but they do contribute significantly to the trajectory cumulative reward and thus, to the likelihood of trajectories. Such actions in states may actually be unsafe. 
    
\end{itemize}

Therefore, considering the problem at the trajectory-level, the agent does not have the flexibility to choose individual actions that would allow it to maintain safety, increasing the likelihood of trajectories in terms of safety, even if some of the states crossed are potentially unsafe.

\section{Problem specification} \label{sec:preb_spec}
To address the above mentioned issues of a trajectory-level inverse constrained reinforcement learning approach, we aim at maximizing the probability of demonstrated trajectories by considering the $Q$ values of individual state-action pairs ($Q: S \times A \rightarrow \mathbb{R}$) comprising these trajectories, mixing the expectation regarding the performance of actions in states on rewards and safety. For the simplicity of presentation, with an abuse of notation, we subsequently use $\mathcal{P}^E$ to denote the distribution of demonstrated steps, i.e., state-action pairs, as well as the distribution of states crossed by any $\tau \in \mathcal{D}_E$. Similarly, we denote $supp^E$ the support set of state-action pairs, as well as the support set of states in any $\tau \in \mathcal{D}_E$.

To formulate the Q function and the final objective, we distinguish two types of reward values: The task-specific reward values (denoted $r_d$) provided by the environment per step (i.e., $r_d: S \times A \rightarrow \mathbb{R}$), and the safety rewards (denoted $r_s$) on states (i.e., $r_s: S  \rightarrow \mathbb{R}$). The latter type of reward should be considered as a ``constraint-abiding" bonus or as a penalty for being in an unsafe state. Without loss of generality, we can assume that for any $(s, a) \in S \times A$, it holds that $r_s(s) \leq r_d(s, a)$. For instance, we can assume, without loss of generality, that $r_s(\cdot) \leq 0$ and $r_d(\cdot) \geq 0$. 

Now, the Q function is defined to be the expected sum of rewards when the agent performs action $a_t$ in state $s_t$ and in subsequent steps it acts according to the policy $\pi$, given the environment dynamics determined by $P$:
\begin{equation} \label{eq:exp_sum_of_rew}
    Q^\pi(s_t,a_t)=\mathbb{E}_{\pi, P}\sum_{i=0}^{T-t-1}\gamma^iR_{t+i}
\end{equation}
where, \[R_t=[\mathbb{I}^{S}(s_{t})r_d(s_{t},a_{t})+(1-\mathbb{I}^{S}(s_{t}))r_s(s_{t})]\]

\noindent and $\mathbb{I}^{S}(s_{t})$ specifies an assessment of whether the state $s_t$ is safe. This can be a binary assessment of whether $(s, a) \in supp^E$ \footnote{It must be noted that a binary $\mathbb{I}^{S}$ does not necessarily assess whether the state is in the $supp^E$.}. 

Given the above specification, the Q value of a state-action pair is an accumulation of a mixture of discounted  ``task specific" and ``safety" rewards, where the latter are action-independent.   
This specifies the safety of a state and the promise of an action $a_t$ performed in state $s_t$ in terms of rewards. To see this, let us consider that starting from $(s_t,a_t)$, a trajectory crosses only unsafe states. Then, $\mathbb{I}^{S}(s_{i})=0, \forall i \in \{t, t+1, ...\}$, and the $Q(s_t,a_t)$ value accumulates only penalties for these states. On the other hand, in case at some point $t' \geq t$ the agent performs an action that recovers safety and then it continues using a policy that maintains safety, then the Q value from that point and on accumulates task specific positive rewards. 
Based on this, we can prove the following theorem:

\textit{Theorem}: When for any $(s,a) \in S \times A$ with $(s,a) \notin supp^E$ it holds that $r_s(s) \leq 0$, and $\mathbb{I}^S$ is binary,   then the following holds for any policy $\pi$:
\begin{align}
    Q^\pi(s,a) \leq  min_{(s^d,a^d) \in supp^E}\{Q^{\pi_d}(s^d,a^d)\}
\end{align}
\noindent where $s^d \in supp^E$ are states that are the closest ones (based on a measure of proximity) to  $s \notin supp^E$ that the agent can reach in subsequent steps, and $Q^{\pi_d}$ are the Q values under the demonstrated policy. It must be noted that this inequality holds in expectation according to the policy $\pi$ and environment dynamics $P$. 

\textit{Proof}. To prove this, let $s_t$ be a state not in $supp^E$. The policy $\pi$, starting from any state, does not guarantee that it results in trajectories that cross only safe states. Since $\mathbb{I}^S$ is binary and $s_t \notin supp^E$, then $\mathbb{I}^S(s_t) =0$ and  $R_t=r_s(s_t)$. Thus, it holds that:
\[Q^\pi(s_t,a_t)=\mathbb{E}_{\pi, P}[r_s(s_t)+\sum_{i=1}^{T-t-1}\gamma^iR_{t+i}]\]
In case the demonstrated policy, $\pi_d$, starts from any state $s^d_t$ where it applies $a^d_t \in A$, s.t. $(s^d_t,a^d_t) \in supp^E$, then 
\[Q^{\pi_d}(s^d_t,a^d_t)= \mathbb{E}_{\pi^*, P} [r_d(s^d_t,a^d_t)+\sum_{i=1}^{T-t-1}\gamma^iR'_{t+i}]\]

\noindent Since $(s^d,a^d) \in supp^E$, then $\mathbb{I}^S(s^d,a^d)=1$ and it holds that  $R'_t=r_d(s^d_t,a^d_t)$. 
 Now, the assumption that for any state $s$, $r_s(s) \leq 0$, implies that $R_t \leq r_d(s^d_t,a^d_t)$, for any $(s^d_t,a^d_t) \in supp^E$. Hence, $R_t \leq min_{(s^d,a^d) \in supp^E} r_d(s^d_t,a^d_t)$. This is the case for any subsequent state not in $supp^E$ visited by $\pi$ starting from $s_t$ in comparison to states visited by the demonstrated policy. For states in $supp^E$ visited by $\pi$, their Q values cannot be greater than those visited by the policy ${\pi_d}$, i.e., for $s_t^d \in supp^E$, it holds that 
$Q^\pi(s_t^d,\pi(s_t^d)) \leq Q^{\pi_d}(s_t^d,a_t^d)$. Therefore, the following holds, since, $\pi_d$ crosses only safe states, while $\pi$ may cross unsafe states:
\begin{align*}
    & Q^\pi(s_t,a_t) \leq min_{(s^d,a^d) \in supp^E}(r_d(s^d, a^d))+ \mathbb{E}_{\pi,P}[\gamma Q^\pi(s_{t+1},a_{t+1})] \\ & \leq min_{(s^d,a^d) \in supp^E}\{Q^{\pi_d}(s^d,a^d)\}
\end{align*}

Inequality (3) formulates the idea that state-action pairs known to be safe should have the highest promise in terms of rewards (and safety). Unfortunately, this inequality is not guaranteed to hold in case $\mathbb{I}^S$ is probabilistic.  Therefore, we need to (a) maximize the Q values of state-action pairs that are in the support of demonstrations (i.e., known to be safe), and to (b) maintain the Q values of actions in states that are not in the support of demonstrated states, lower than those in $supp^E$, \emph{in expectation} to the policy, environment dynamics and the uncertainty about the safety of states. 
In so doing, we aim to prevent trajectories not in the demonstrations from being highly probable, and support agents to find actions that, when executed in probably unsafe states, can recover safety with high probability, with respect to the environment dynamics. 

We can now formulate the objective of the inverse constrained reinforcement learning in terms of Q values. In stochastic, continuous domains with respect to the dynamics $P$, we know that we need to maximize: \[\sum_t\mathbb{E}_{\pi,P}  [R_t+\mathcal{H}(\pi(a_t|s_t))],\] 
\noindent where the expectation is in the distribution of state-actions under the policy $\pi$ and environment dynamics. This is optimized by choosing $\pi(a_t|s_t)=exp(Q(s_t,a_t)-V(s_t))$, with a soft maximization of the value function: $V(s_t)=log\int exp(Q(s_t,a_t))da_t$. To evaluate the policy according to the maximum entropy objective, given a policy $\pi$, we update $Q$ values following the Bellman update: 
\[Q(s_t,a_t)=R_t + \gamma \mathbb{E}_{s_{t+1} \sim P}V(s_{t+1})\]
\noindent where the soft value function is as follows:
\[V(s_t)=\mathbb{E}_{a_t \sim \pi}[Q(s_t,a_t)-log\pi(a_t|s_t)],\]
\noindent transforming the policy entropy specified in (1) to a causal entropy on individual steps.

Therefore, considering that $\pi(a_t|s_t) \propto exp(Q(s_t,a_t))$, for the joint probability of actions ($CQ$)  in a trajectory $\tau$ it holds that:
\[CQ(\tau) \propto \prod_{t=0}^T exp(Q(s_t,a_t)),\]

\noindent Thus, it holds that:
\begin{align*} 
& logp(\mathcal{D}_E) \propto  log[\prod_{i=1}^{|\mathcal{D}_E|}(\prod_{t=0}^T exp(Q(s^i_t,a^i_t)]
\end{align*}
\noindent We aim to maximize the likelihood of demonstrated trajectories by maximizing the following term:
\begin{align} 
& max [\sum_{i=1}^{|\mathcal{D}_E|}\sum_{t=0}^T Q(s^i_t,a^i_t)]
\end{align}
\noindent subject to: 
\begin{align} \label{eq:q-constraint}
    Q(s,a) \preceq min_{(s',a') \in supp^E}\{Q(s',a')\}
\end{align}
\noindent for any $(s,a) \in S \times A$ with $s \notin supp^E$. 
The symbol $\preceq$ denotes that the inequality (\ref{eq:q-constraint}) holds in expectation, depending on the assessment $\mathbb{I}^S$ on states, the policy $\pi$, and the environment dynamics $P$. This does not prevent pairs $(s,a) \notin \mathcal{P}^E$ from having Q values that are greater than the minimum Q value of the demonstrated steps. 


\section{Safe Q-Learning}
\subsection{Objective function}

Guided by the max-entropy formulation and the step-wise likelihood view in Section \ref{sec:preb_spec}, we explicitly enforce the constraint (\ref{eq:q-constraint}) in expectation on states $s \notin \mathcal{P}^E$, and update Q-values on states in or out of $\mathcal{P}^E$. In so doing, we allow the agent to learn both from demonstrations and online interactions. Concretely, let $B$ be a buffer of policy rollouts and $D$ the demonstration buffer. 

Getting samples from $B$, these may be in $\mathcal{P}^E$ or not. In the latter case, we need to enforce the constraint (\ref{eq:q-constraint}). 
To do this, for each state $s_B$ in a sample $(s_B, a_B,r_{d}(s_B, a_B),s_B',a_B')$ from $B$, we search and pick a sample from $D$ with the ``closest" state $s^*_D$ (we elaborate on this in Section \ref{sec:algo}), together with its accompanying action $a_D^*$, task reward $r_{d}^*(s^*_D, a^*_D)$, and next state $s_{D}'^{*}$. In addition, we get the next action $a_{D}'^{*}$ and use the estimated Q-target value of $(s_D'^*, a_D'^*)$, 
\[\hat{Q}_{min}(s_B,a_B)=r_{d}^*(s^*_D, a^*_D) + \gamma Q(s_D'^*, a_D'^*),\] 
\noindent as a local bound target for $Q(s_B,a_B)$\footnote{In this section, we deliberately omit the entropy term. In addition, we omit details on how exactly $s_D^*$, $a_D^*$, $r_{d, D}^*$, $s_D'^*$, $a_D'^*$ are found, and whether we use the next action $a'_D$ from the buffer or from the current policy (as, for instance, in SAC). These details are included in Section \ref{sec:algo}.}. Then, the objective for samples from $B$ is as follows:
\begin{align} \label{eq:L^B_Q}
    \mathcal{L}^B_Q = \sum_{(s_B,a_B,r_{d,B},s_B',a_B') \sim B} &[(1-\mathbb{I}^S(s_B)) \cdot LC^B_{s_B \notin \mathcal{P}^E}(s_B,a_B) \nonumber \\
    &+   \mathbb{I}^S(s_B) \cdot L^B_{s_B \in \mathcal{P}^E}(s_B,a_B,r_{d,B},s_B',a_B')]  
\end{align}
where:
\begin{align*}
    &LC^B_{s_B \notin \mathcal{P}^E}(s_B,a_B) = \\
    & \ \ \ \ \ \left( max\left( Q(s_B,a_B), \hat{Q}_{min}(s_B,a_B) \right) - \hat{Q}_{min}(s_B,a_B) \right)^2 
\end{align*}
and: 
\begin{align*}
    L^B_{s_B \in \mathcal{P}^E}&(s_B,a_B,r_{d,B},s_B',a_B') =\\ &\left( Q(s_B, a_B) - [r_{d}(s_B, a_B) + \gamma Q(s_B',a_B')] \right)^2
\end{align*}

\noindent where the first term in brackets in Equation \ref{eq:L^B_Q} enforces the constraint. However, we also have to include the safety reward term in case $s_B \notin \mathcal{P}^E$:
\begin{equation*}
    L^B_{s_B \notin \mathcal{P}^E}(s_B,a_B,s_B',a_B') = \left( Q(s_B,a_B) - [r_s(s_{B}) + \gamma Q(s_B',a_B')] \right)^2
\end{equation*}

 So, now the objective becomes:
\begin{align}
    \mathcal{L}^B_Q = \sum_{(s_B,a_B,r_{d,B},s_B',a_B') \sim B} &[(1-\mathbb{I}^S(s_B)) \cdot LC^B_{s_B \notin \mathcal{P}^E}(s_B,a_B) \nonumber \\
    &+ (1-\mathbb{I}^S(s_B)) \cdot L^B_{s_B \notin \mathcal{P}^E}(s_B,a_B,s_B',a_B') \nonumber \\
    &+ \mathbb{I}^S(s_B) \cdot L^B_{s_B \in \mathcal{P}^E}(s_B,a_B,r_{d,B},s_B',a_B')]  
\end{align}

To recap, the above objective covers: (a) the task-specific reward term and the safety reward term, both weighted by the safety estimation, as proposed in Equation \ref{eq:exp_sum_of_rew}, and (b) it enforces the constraint of Equation \ref{eq:q-constraint}. 
Based on previous work \cite{hester2018deep} and on our preliminary experiments, we found it beneficial to introduce a bias in Q-values towards the demonstrations' expected reward. Therefore, we incorporate the term $\mathcal{L}^D_Q$ that explicitly updates the Q-values based on samples from $D$ comprising the demonstrated state-action pairs $(s_D,a_D)$, the next state-action pairs $(s_D',a_D')$, and the corresponding received task reward $r_{d}(s_D,a_D)$:

\begin{align*}
\mathcal{L}^D_Q=\sum_{(s_D, a_D,r_{d,D},s_D',a_D') \sim D} L^D(s_D,a_D, r_{d}(s_D,a_D), s_D',a_D')
\end{align*}
where:
\begin{align*}
    &L^D(s_D,a_D, r_{d}(s_D,a_D), s_D',a_D') = \\ &\left( Q(s_D,a_D) - [r_{d}(s_D,a_D) + \gamma Q(s_D', a_D')] \right)^2
\end{align*}

\noindent This term is not weighted by the discriminator's estimate since the samples come from the demonstrations' distribution. So, the final objective, given $N$ samples from $B$ and $N$ samples from $D$, is formulated as follows:
\begin{align} \label{eq:final_obj}
    \mathcal{L}_Q = \frac{1}{2N} [\mathcal{L}^B_Q + \mathcal{L}^D_Q] 
\end{align}

\subsection{The SafeQIL algorithm} \label{sec:algo}

In this section, we provide a practical algorithm using the proposed objective and SAC as the backbone algorithm. 

We maintain a stochastic policy (actor) $\pi_\theta(a|s)$, two critics $Q_{\phi_1}$, $Q_{\phi_2}$ and two target critics $Q_{\bar{\phi}_1}$, $Q_{\bar{\phi}_2}$ as well as, a discriminator $\phi_\omega$, that estimates the probability for a state to be in $\mathcal{P}^E$. Also, the safety reward/penalty $r_s(s)$ is calculated based on the discriminator's estimate $\phi_\omega(s)$. Specifically, we choose $r_s(s)=log \left( \phi_\omega(s) \right)$ which maps the discriminator's estimate (range $[0,1]$) to a negative safety reward (range $[-\infty, 0]$).

Online rollouts are stored in $B$ and demonstrations in $D$. During the update of Q networks (critics), we get $N$ samples from $B$ and $N$ samples from $D$, and each critic is updated separately using Equation \ref{eq:final_obj}. 
An important detail about Equation \ref{eq:final_obj}  requiring clarification is about how we retrieve ($s_D^*$, $a_D^*$, $r_D^*$, $s_D'^*$,$a_D'^*$) from $D$ given a state $s_B$. Actually, we get the index of the sample $s_D^*$ that minimizes the cosine similarity to $s_B$:
\begin{equation}\label{eq:argmin_cossim}
    i_D^* = \arg\min \left( \text{CosSim}(\mathbf{s_B}, \mathbf{S_D}) \right)
\end{equation}
Having the index $i_D^*$, it is straightforward to slice buffer $D$ and get the corresponding sample for the calculation of $\hat{Q}_{min}$.

Continuing on the details about the entropy and Q-networks weights, assuming that we update critic $j$:
\begin{align}\label{eq:final_constr-term}
        &LC^B_{j, s_B \notin \mathcal{P}^E}(s_B,a_B) = \\
    & \ \ \ \ \ \left( max\left( Q_{\phi_j}(s_B,a_B), \hat{Q}_{min}(s_B,a_B) \right) - \hat{Q}_{min}(s_B,a_B) \right)^2  \nonumber 
\end{align}
where (with an abuse of notation for simplicity of presentation):
\begin{equation*}
    \hat{Q}_{min}(s_B,a_B) = r_{d}^*(s_B,a_B) + \gamma \cdot \min_{i=1,2} Q_{\bar{\phi}_i}(s_D'^*, a_D'^*)
\end{equation*}

The next action $a'$ is predicted based on the current policy for $L^B_{s_B \notin \mathcal{P}^E}(s_B,a_B,r_{d,B},s_B',a_B')$ (i.e., the term for states out of $\mathcal{P}^E$), while for $L^D(s_D,a_D, r_{d}(s_D,a_D), s_D',a_D')$ (i.e., the term for states in $\mathcal{P}^E$) we use the one stored in $D$. In so doing, we omit the entropy term in both terms $L^B_{s_B \notin \mathcal{P}^E}(s_B,a_B,r_{d,B},s_B',a_B')$ and  $L^D(s_D,a_D, r_{d}(s_D,a_D), s_D',a_D')$ aligned with $LC^B_{j, s_B \notin \mathcal{P}^E}(s_B,a_B)$. These terms for the critic $j$ can be expressed as follows:
\begin{align}\label{eq:final_ood-term}
&L^B_{j, s_B \notin \mathcal{P}^E}(s_B,a_B,s_B',a_B') =\\ \nonumber &\left( Q_{\phi_j}(s_B,a_B) - [r_s(s_B)+ \gamma \cdot \min_{i=1,2}Q_{\bar{\phi_i}, \ a' \sim \pi_{\theta}(\cdot|s_B')}(s_B',a')] \right)^2
\end{align}
and
\begin{align}\label{eq:final_dem-bias-term}
    &L^D_j(s_D,a_D, r_{d}(s_D,a_D), s_D',a_D') =\\ \nonumber  &\left( Q_{\phi_j}(s_D,a_D) - [r_{d}(s_D,a_D) +\gamma \cdot \min_{i=1,2} Q_{\bar{\phi}_i}(s_D', a_D')]\right)
\end{align}

Regarding $L^B_{s_B \in \mathcal{P}^E}(s_B,a_B,r_{d,B},s_B',a_B')$, we don't perform any change to the backbone SAC algorithm, to maintain its strengths, so this term for the critic $j$  is expressed as follows:
\begin{align}\label{eq:final_sac-term}
    & L^B_{j, s_B \in \mathcal{P}^E}(s_B,a_B,r_{d,B},s_B',a_B') = \\ \nonumber
     (  & Q_{\bar{\phi}_j}(s_B, a_B) - \\ &[r_{d}(s_B, a_B) + \gamma \left( \min_{i=1,2} Q_{\bar{\phi_i}, \ a' \sim \pi_{\theta}(\cdot|s_B')}(s_B',a') -\alpha \log\pi_\theta(a'|s_B') \right)])^2 \nonumber
\end{align}

\noindent where $\alpha$ is the entropy coefficient. To update the policy, we maximize the soft objective:
\begin{equation} \label{eq:policy_obj}
    J_{\pi_\theta} = \frac{1}{N}\sum_{s_B \sim B} \left( \min_{i=1,2} Q_{\bar{\phi_i}, \ a \sim \pi_{\theta}(\cdot|s_B)}(s_B,a) -\alpha \log\pi_\theta(a|s_B) \right)
\end{equation}
\noindent while the auto-tuned entropy coefficient is updated using the loss defined in \cite{haarnoja2018soft}:
\begin{equation}\label{eq:entr-coef_obj}
    \mathcal{L}_{\alpha} = \frac{1}{N}\sum_{s_B \sim B} \left( - \left( \alpha \log\pi_\theta(a|s_D) + \bar{\mathcal{H}} \right) \right)
\end{equation}
\noindent where $\bar{\mathcal{H}}$ is a hyperparameter. Each target critic $j$ is updated using the Polyak averaging with the corresponding parameter $\eta$:
\begin{equation}\label{eq:target-critic_upd}
    \bar{\phi}_j \leftarrow \eta \bar{\phi}_j + (1-\eta) \phi_j
\end{equation}

Finally, at each update of all the aforementioned, we also update the discriminator using the Logistic Loss as used in DAC \cite{kostrikov2018discriminatoractorcritic}, the Gradient Penalty regularization of \cite{gulrajani2017improved}, $N$ samples from $B$ and $N$ samples from $D$:
\begin{align} \label{eq:discr_loss}
    &\mathcal{L}_\omega = \\
    &\frac{1}{2N} \left( \sum_{s_B \sim B} \left[ -\log \left(1-\phi_{\omega}(s_B) \right) \right] + \sum_{s_D \sim D} \left[ -\log \left(\phi_{\omega}(s_D) \right) \right] \right) \nonumber \\
    &+ \lambda_{\mathrm{gp}} \cdot \frac{1}{2N}
   \sum_{(s_B,\,s_D)} \left(\left\lVert\nabla_{\hat{s}}\phi_{\omega}(\hat{s})\right\rVert_2 - 1\right)^2 \nonumber
\end{align}
where:
\begin{equation*}
    \hat{s} \;=\; \epsilon\, s_D \;+\; (1-\epsilon)\, s_B, \qquad \epsilon \sim \mathcal{U}(0,1).
\end{equation*}
We summarize all the above in Algorithm \ref{alg:SafeQIL}. In summary, the method couples a support-aware constraint with max-entropy policy learning: demonstrations provide local upper bounds for the value function of state-action pairs not in the distribution of demonstrations, the discriminator softly gates updates so that online interactions improve performance on in-distribution states, while a safety reward drives recovery off-distribution. This yields a modification of standard soft actor–critic training that curbs unsafe extrapolation, preserves performance in states in the support of demonstrations, and, in special cases, it can be reduced to SAC if all states are in-distribution of demonstrations.

\begin{algorithm}[t]
\caption{SafeQIL}
\label{alg:SafeQIL}
\begin{algorithmic}[1]

\Require Demonstration buffer \(D\)
\State Initialize Policy \(\pi_\theta\); Critics \(Q_{\phi_1},Q_{\phi_2}\); Target critics \(Q_{\bar\phi_1}, Q_{\bar\phi_2}\); Entropy coefficient \(\alpha\); Discriminator \(\phi_\omega\); Replay buffer \(B\gets\varnothing\)

\For{each environment step}
  \State Observe \(s_t\); Sample and apply \(a_t\sim\pi_\theta(\cdot|s_t)\); Get \((r_{d,t}, s_{t+1})\)
  \State Push \((s_t,a_t,r_{d,t},s_{t+1})\) into \(B\)
  \For{update step \(u=1,\dots,U\)}
    \State Sample \(N\) tuples \((s_B,a_B,r_{d,B},s_B')\sim B\) 
    \State Sample \(N\) tuples \((s_D,a_D,r_{d,D},s_D')\sim D\)
    \State For each \(s_B\) get index \(i_D^*\) \hfill (Eq. \ref{eq:argmin_cossim})
    \State For each \(i_D^*\) retrieve \((r_{d,D}^*,s_D'^{*},a_D'^*)\) slicing \(D\)
    \State Sample \(a_B'\sim\pi_\theta(\cdot|s_B')\)
    \State Update discriminator \(\phi_\omega\) \hfill (Eq. \ref{eq:discr_loss})
    \State Update critics \(Q_{\phi_1},Q_{\phi_2}\) \hfill (Eq. \ref{eq:final_obj}, \ref{eq:final_constr-term}, \ref{eq:final_ood-term}, \ref{eq:final_dem-bias-term}, \ref{eq:final_sac-term})
    \State Update policy \(\pi_\theta\) \hfill (Eq. \ref{eq:policy_obj}) 
    \State Update entropy coefficient \(\alpha\) \hfill (Eq. \ref{eq:entr-coef_obj})
    \State Update target critics \(Q_{\bar\phi_1}, Q_{\bar\phi_2}\) \hfill (Eq. \ref{eq:target-critic_upd})
  \EndFor
\EndFor
\end{algorithmic}
\end{algorithm}


\section{Experiments}

\subsection{Experimental settings}\label{sec:exp_set}

\begin{figure}[h]
  \centering
  \includegraphics[width=0.32\textwidth]{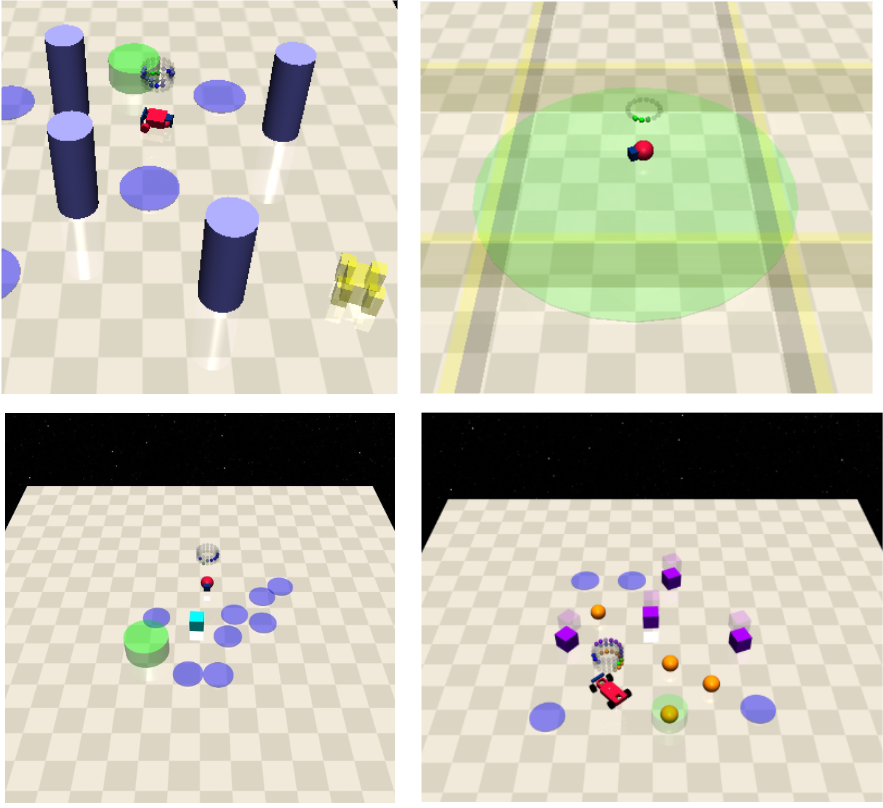}
  
  \caption{Snapshots of: (top-left) SafetyCarPush2‑v0, (top-right) SafetyPointCircle2-v0, (bottom-left) SafetyPointGoal1-v0, (bottom-right) SafetyCarButton1-v0.}
  \label{fig:benchs}
\end{figure}

\textbf{Tasks.} We evaluate SafeQIL on 4 tasks from Safety-Gymnasium: SafetyPointGoal1-v0, SafetyPointCircle2-v0, SafetyCarButton1-v0, and SafetyCarPush2-v0, covering navigation and object interaction under safety constraints. Figure \ref{fig:benchs} provides representative snapshots of the 4 experimental settings to illustrate task geometry and typical safety-related objects. This set of tasks spans increasing difficulty and safety density, that is, from boundary-only constraints (SafetyPointCircle2-v0) to multi-obstacle navigation (SafetyPointGoal1-v0) and interaction-heavy control (SafetyCarButton1-v0/SafetyCarPush2-v0), providing a broad stress-test for algorithms that must (i) improve in-distribution while (ii) remaining conservative at out-of-distribution states.

\textbf{Demonstrations.} To train the algorithms, we collect human-generated demonstration datasets for each task via keyboard control, producing (state, action, reward, cost, next state, next action) tuples. To the best of our knowledge, existing offline safe-RL benchmarks \cite{liu2023datasets} provide algorithm-generated datasets across 38 Safety-Gymnasium / BulletSafetyGym / MetaDrive tasks, but there is no publicly available human-generated demonstration set for Safety-Gymnasium tasks. 

\textbf{Baselines.} We compare SafeQIL against 3 strong baselines:

- ICRL \cite{pmlr-v139-malik21a} learns an explicit constraint function from demonstrations and then optimizes a policy under the inferred constraints. Our implementation relies on the official ICRL implementation \cite{malik2021icrl_code}.

- VICRL \cite{liu2023benchmarking} models a posterior distribution over constraints via variational inference, and our implementation is based on the official VICRL implementation \cite{liu2023icrl_benchmarks}.

- SAC-GAIL: a GAIL-style discriminator provides the reward while the actor and critic are trained with SAC. We implement this in the spirit of off-policy AIL, such as DAC \cite{kostrikov2018discriminatoractorcriticaddressingsampleinefficiency}, which couples a GAIL discriminator with an off-policy actor-critic. This baseline tests whether pure imitation suffices on our tasks, as opposed to SafeQIL’s value-shaping with state-level pessimism.

ICRL and VICRL were extensively tuned using the complete set of configurations from their original papers/implementations (3 and 9 settings, respectively) on SafetyPointGoal1-v0. Because the best ICRL configuration (Setting 2) involves a low number of environment steps, we also evaluate ICRL using the best VICRL parameters (Setting 7), tuning the regularization coefficient per task for both configurations. We report results for both, denoted as ICRL$^2$ and ICRL$^7$. For SafeQIL and SAC-GAIL, we utilized the Stable Baselines3 \cite{raffin2021stable} SAC implementation. We use default hyperparameters for these baselines, tuning only the regularization coefficient per task. All the hyperparameter settings are provided in the Appendix \ref{app:hyperparameters}.

Finally, we include results from the unconstrained backbone algorithms (SAC and PPO), as well as the Human Demonstrator, to provide a comprehensive comparison.

\textbf{Evaluation protocol.} For each task, every method is trained with the same set of 40 human-generated demonstration trajectories. After training, we run 40 evaluation episodes per method. We measure episodic reward (higher is better) and safety cost (lower is better) on the 4 Safety-Gymnasium tasks. We repeat this with 3 independent random seeds and report mean and standard deviation (in the form mean ± std) over seeds, following recommended reporting practices for reliability.

\subsection{Experimental results}\label{sec:exp_res}

Tables \ref{tab:perf-pointgoal}–\ref{tab:perf-carpush} and Figure \ref{fig:pointgoal_exp}, as well as Figures \ref{fig:pointcircle_exp}-\ref{fig:carpush_exp} in Appendix \ref{app:more_curves} summarize results. The best algorithm is decided (highlighted) based on the following process: If all the algorithms exceed the unconstrained SAC cost, the one with the minimum cost is selected. Otherwise, we highlight the algorithm with the best trade-off between cost reduction and reward performance relative to the baseline. Hyperparameters of SafeQIL and baseline algorithms have been tuned towards improving this trade-off. The detailed methodology for quantifying the trade-off is described in Appendix \ref{app:quant_analysis} along with detailed results.

\textbf{SafetyPointGoal1-v0.}  In this navigation task, the unconstrained SAC baseline incurs a cost of $49.15 \pm 2.21$. Surprisingly, the inverse constraint baselines fail to improve upon this baseline, with ICRL and VICRL incurring higher costs ($62.60$ and $62.97$, respectively). SafeQIL emerges as the best algorithm, achieving a cost of $34.22 \pm 2.71$, which corresponds to a 30.4$\%$ reduction in cost relative to the unconstrained SAC baseline. While SAC-GAIL also reduces the cost (to $44.80$), SafeQIL’s reduction is significantly deeper. Although SafeQIL’s reward ($5.27$) is lower than the unconstrained baselines, it is the only method that improves safety margins over the baseline while maintaining positive task performance.

\begin{table}[ht]
\caption{SafetyPointGoal1-v0: Comparative results.}
\label{tab:perf-pointgoal}
\centering
\begin{tabular}{lcc}
\toprule
Algorithm & Reward~$\uparrow$ & Cost~$\downarrow$ \\
\midrule
\textbf{SafeQIL} & \textbf{5.27 $\pm$ 1.85} & \textbf{34.22 $\pm$ 2.71} \\ 
ICRL$^2$ & 23.21 $\pm$ 1.28 & 62.60 $\pm$ 9.87 \\
ICRL$^7$ & -0.50 $\pm$ 4.93 & 42.80 $\pm$ 35.43 \\
VICRL & 21.87 $\pm$ 1.20 & 62.97 $\pm$ 10.81 \\
SAC-GAIL & 7.17 $\pm$ 1.65 & 44.80 $\pm$ 18.60 \\
\midrule
SAC & 27.47 $\pm$ 0.21 & 49.15 $\pm$ 2.21 \\
PPO$^2$ & 23.64 $\pm$ 1.64 & 60.99 $\pm$ 7.99 \\
PPO$^7$ & 26.70 $\pm$ 0.11 & 57.17 $\pm$ 1.68 \\
Human Demonstrator & 11.39 $\pm$ 1.55 & 0.00 $\pm$ 0.00 \\
\bottomrule
\end{tabular}
\end{table}

\begin{figure}[h]
  \centering
  \includegraphics[width=0.46\textwidth]{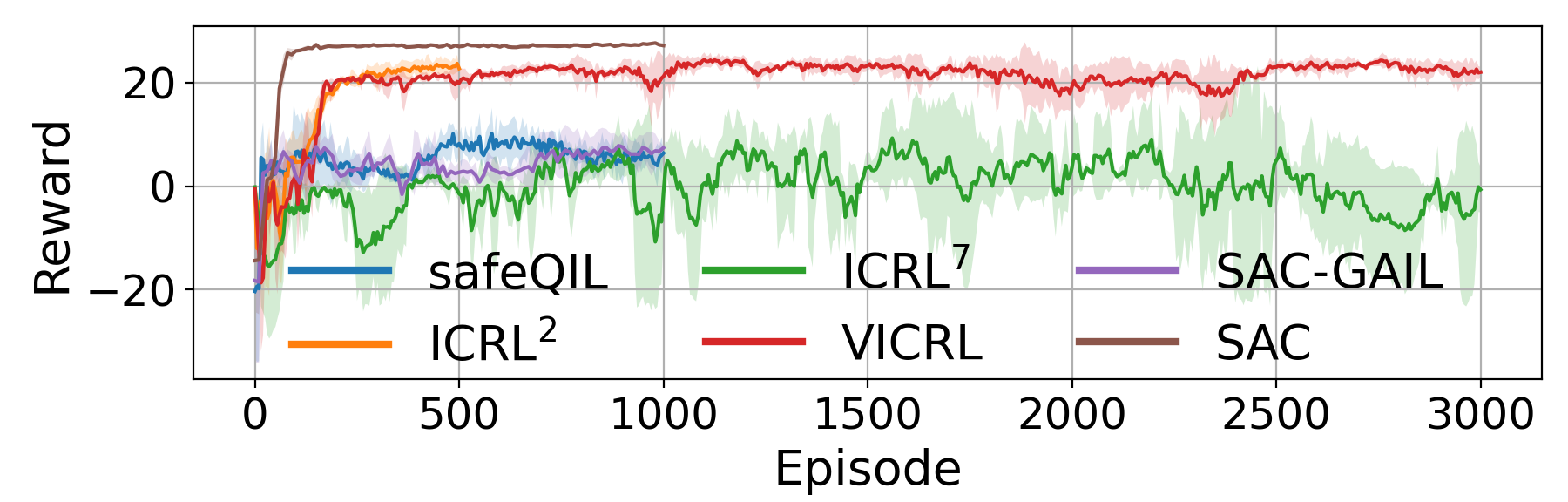}
  
  \includegraphics[width=0.45\textwidth]{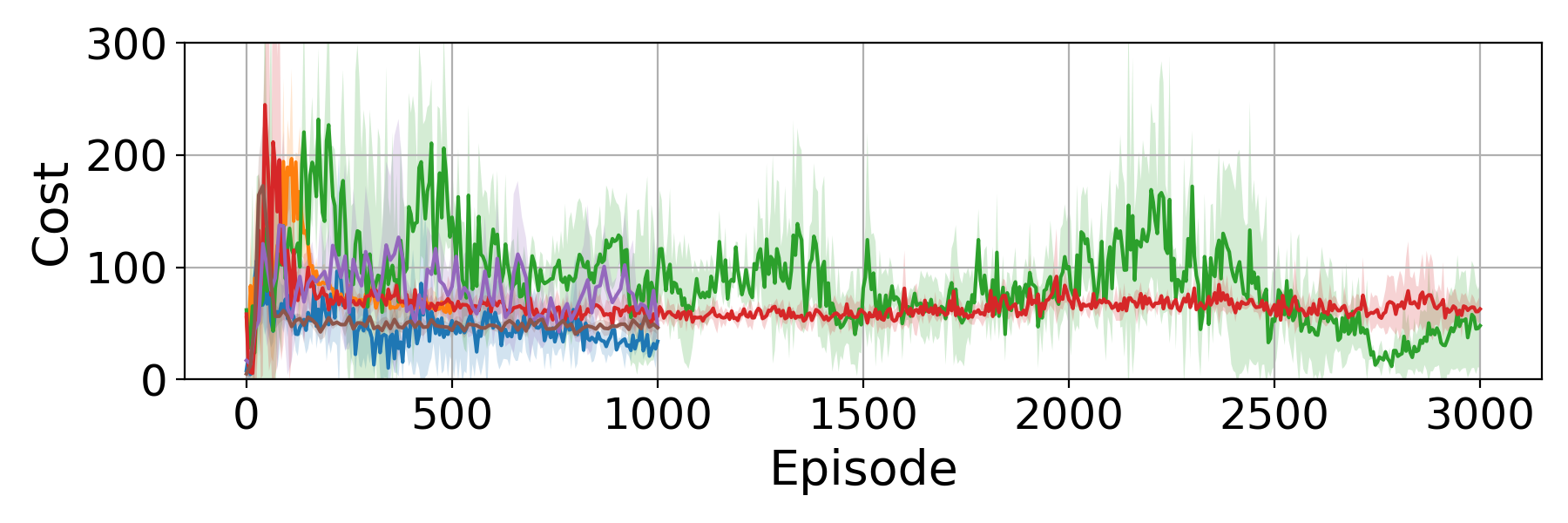}
  
  \caption{Learning curves for SafetyPointGoal1-v0.}
  \label{fig:pointgoal_exp}
\end{figure}

\textbf{SafetyPointCircle2-v0.} The unconstrained SAC baseline incurs a massive cost of $392.20 \pm 4.38$ in this task. Both SafeQIL and VICRL successfully identify the boundary constraints, significantly outperforming ICRL$^2$. VICRL achieves the best safety performance with a cost of $5.49 \pm 0.81$, representing a 98$\%$ reduction relative to the unconstrained baseline. SafeQIL is very close behind, achieving a cost of $29.28 \pm 8.41$, which corresponds to a 92$\%$ cost reduction. Crucially, SafeQIL offers a far better trade-off compared to ICRL$^7$ and SAC-GAIL: although these baselines achieve similar safety levels (reducing cost by 91$\%$), they suffer severe reward degradation, retaining only 14$\%$ and 20$\%$ of the baseline reward, respectively. In contrast, SafeQIL retains 46$\%$ of the baseline reward, effectively matching VICRL's performance, outperforming the other constrained baselines by more than double in terms of task performance.

\begin{table}[h]
\caption{SafetyPointCircle2-v0: Comparative results.}
\label{tab:perf-pointcircle}
\centering
\begin{tabular}{lcc}
\toprule
Algorithm & Reward~$\uparrow$ & Cost~$\downarrow$ \\
\midrule
SafeQIL & 27.06 $\pm$ 4.10 & 29.28 $\pm$ 8.41 \\ 
ICRL$^2$ & 28.36 $\pm$ 17.37 & 177.88 $\pm$ 6.70 \\
ICRL$^7$ & 8.21 $\pm$ 17.07 & 34.07 $\pm$ 16.74 \\
\textbf{VICRL} & \textbf{26.29 $\pm$ 0.69} & \textbf{5.49 $\pm$ 0.81} \\
SAC-GAIL & 11.89 $\pm$ 9.55 & 33.95 $\pm$ 28.02 \\
\midrule
SAC & 58.81 $\pm$ 0.70 & 392.20 $\pm$ 4.38  \\
PPO$^2$ & 43.24 $\pm$ 12.01 & 250.40 $\pm$ 217.83 \\
PPO$^7$ & 32.57 $\pm$ 1.26 & 408.74 $\pm$ 2.76 \\
Human Demonstrator & 24.28 $\pm$ 3.98 & 0.00 $\pm$ 0.00 \\
\bottomrule
\end{tabular}
\end{table}

\textbf{SafetyCarButton1-v0.} This interaction-heavy task proves difficult for all algorithms. The unconstrained SAC baseline achieves a high reward but incurs a high cost of $299.27 \pm 20.61$. SafeQIL effectively bridges the gap, reducing the cost to $70.11 \pm 72.96$, which represents a 76$\%$ safety improvement over the baseline. While VICRL achieves an even higher cost reduction of 88$\%$ ($34.65 \pm 23.36$), it suffers from a complete collapse in task performance (Reward $=-14.12$), failing to solve the task. SafeQIL avoids this collapse (Reward $=-3.81$), offering the most efficient trade-off by maintaining a performing policy that enforces safety.

\begin{table}[h]
\caption{SafetyCarButton1-v0: Comparative results.}
\label{tab:perf-carbutton}
\centering
\begin{tabular}{lcc}
\toprule
Algorithm & Reward~$\uparrow$ & Cost~$\downarrow$ \\
\midrule
\textbf{SafeQIL} & \textbf{-3.81 $\pm$ 3.05} & \textbf{70.11 $\pm$ 72.96} \\ 
ICRL$^2$ & 0.31 $\pm$ 2.23 & 222.01 $\pm$ 173.61 \\
ICRL$^7$ & -15.83 $\pm$ 12.98 & 145.94 $\pm$ 23.54 \\
VICRL & -14.12 $\pm$ 10.83 & 34.65 $\pm$ 23.36 \\
SAC-GAIL & 0.14 $\pm$ 4.72 & 98.15 $\pm$ 81.54 \\
\midrule
SAC & 19.49 $\pm$ 3.50 & 299.27 $\pm$ 20.61 \\
PPO$^2$ & 11.70 $\pm$ 1.90 & 460.47 $\pm$ 6.61 \\
PPO$^7$ & 11.71 $\pm$ 1.16 & 431.77 $\pm$ 40.23 \\
Human Demonstrator & 14.11 $\pm$ 5.67 & 0.00 $\pm$ 0.00 \\
\bottomrule
\end{tabular}
\end{table}

\textbf{SafetyCarPush2-v0.} In this manipulation task, similarly to the Button task, the unconstrained SAC baseline incurs a high cost of $245.36 \pm 28.52$. A critical comparison arises between SafeQIL and SAC-GAIL. SAC-GAIL achieves the most efficient trade-off, reducing the cost by 64$\%$ ($87.97 \pm 10.57$) while maintaining a positive reward of $0.76 \pm 0.92$. However, for applications prioritizing strict safety over task efficiency, SafeQIL demonstrates better performance, lowering the cost further to $57.20 \pm 30.02$ (a 76$\%$ reduction). Although this comes at the expense of a slightly negative reward ($-0.62$), SafeQIL reduces the safety violations by  30 more points compared to SAC-GAIL, providing a much tighter safety bound. In contrast, VICRL matches SafeQIL's cost ($55.80$) but suffers a catastrophic reward drop to $-9.05$, effectively failing the task. Finally, ICRL$^2$ and ICRL$^7$ fail to provide meaningful safety, reducing the cost only by 27$\%$ ($178.81 \pm 60.69$) and 17$\%$ (203.10 $\pm$ 37.24), respectively.

\begin{table}[h]
\caption{SafetyCarPush2-v0: Comparative results.}
\label{tab:perf-carpush}
\centering
\begin{tabular}{lcc}
\toprule
Algorithm & Reward~$\uparrow$ & Cost~$\downarrow$ \\
\midrule
SafeQIL & -0.62 $\pm$ 0.65 & 57.20 $\pm$ 30.02 \\ 
ICRL$^2$ & 0.00 $\pm$ 1.73 & 203.10 $\pm$ 37.24 \\
ICRL$^7$ & 1.15 $\pm$ 0.49 & 178.81 $\pm$ 60.69 \\
VICRL & -9.05 $\pm$ 8.58 & 55.80 $\pm$ 71.45 \\
\textbf{SAC-GAIL} & \textbf{0.76 $\pm$ 0.92} & \textbf{87.97 $\pm$ 10.57} \\
\midrule
SAC & 1.50 $\pm$ 0.01 & 245.36 $\pm$ 28.52 \\
PPO$^2$ & 0.44 $\pm$ 0.52 & 316.50 $\pm$ 52.84 \\
PPO$^7$ & 0.72 $\pm$ 0.64 & 238.63 $\pm$ 23.33 \\
Human Demonstrator & 7.46 $\pm$ 1.77 & 0.00 $\pm$ 0.00 \\
\bottomrule
\end{tabular}
\end{table}

Overall, SafeQIL consistently lowers safety costs relative to the unconstrained baseline, with reductions ranging from 30$\%$ to 92$\%$. It outperforms ICRL in safety across all tasks, and provides a more stable reward-cost trade-off compared to VICRL, which tends to over-constrain the policy to the point of task failure in complex manipulation environments. Finally, compared to SAC-GAIL, SafeQIL demonstrates a more robust safety behavior. While SAC-GAIL offers a competitive baseline in terms of reward, SafeQIL consistently achieves significantly tighter worst-case safety bounds when taking into account also the analysis in Appendix \ref{app:quant_analysis}.


\subsection{Ablation study} \label{sec:abl_study}

To validate the contribution of SafeQIL's components, we conducted an ablation study on the challenging SafetyPointGoal1-v0 benchmark. Table \ref{tab:ablation_main} summarizes the results for the key variations. The original formulation achieves the most effective balance of reward ($5.27 \pm 1.85$) and cost ($34.22 \pm 2.71$). We found that decoupling the upper-bound selection from state similarity ("w/o cosine similarity") results in highly unstable costs ($30.25 \pm 19.10$) and reduced reward, confirming that state-specific bounds are crucial for consistent learning. Similarly, removing the explicit upper-bound constraint ("w/o constraint term") leads to high cost variance ($\pm 14.03$), indicating that relying solely on demonstration bias is insufficient to guarantee consistent safety. For further analysis, Appendix \ref{app:ablation_full} details the full ablation of SafeQIL's components, and Appendix \ref{app:dataset_sensitivity} evaluates the impact of demonstration dataset size.

\begin{table}[h]
\caption{Ablation results for key SafeQIL's variations.}
\label{tab:ablation_main}
\centering
\begin{tabular}{lcc}
\toprule
Algorithm & Reward~$\uparrow$ & Cost~$\downarrow$ \\
\midrule
\textbf{Original} & \textbf{5.27 $\pm$ 1.85} & \textbf{34.22 $\pm$ 2.71} \\
w/o cosine similarity & 3.74 $\pm$ 2.52 & 30.25 $\pm$ 19.10 \\
w/o constraint term & 4.34 $\pm$ 4.41 & 29.02 $\pm$ 14.03 \\
\bottomrule
\end{tabular}
\end{table}


\section{Related work}
This section reviews works that connect most closely to SafeQIL: (i) learning constraints from demonstrations, and (ii) offline RL approaches that enforce pessimism on out-of-distribution (OOD) actions.

\textbf{Learning constraints from demonstrations}. Safety and constraints' abiding behavior is a long-term topic of interest in the RL realm. It was formulated at first by \cite{Altman1999ConstrainedMD} through CMDPs. More recently, the family of Inverse Constraint RL (ICRL) algorithms has shown promising results on learning to respect constraints in settings with an unknown cost function and without cost labels provided. ICRL \cite{pmlr-v139-malik21a} aims to approximate the cost function and employs a binary classifier to differentiate between the demonstrated trajectories and those generated by the policy under learning. A later work formulates constraint inference variationally, introducing a benchmark and a VI-based estimator referred to as VICRL \cite{liu2023benchmarking}. Confidence-aware variants of ICRL \cite{subramanian2024confidenceawareinverseconstrained} explicitly encode a desired confidence level over inferred constraints, seeking estimates that are at least as strict as ground truth with high probability — aiming to address the conservatism/coverage trade-off inherent in constraint inference. Continuing, Multi-Modal ICRL (MMICRL) \cite{NEURIPS2023_bdc48324} relaxes the single-expert assumption by modeling demonstrations as mixtures from heterogeneous experts who may adhere to distinct constraint sets. It infers multiple constraint modes and shows improved recovery and control performance in both discrete and continuous domains. Uncertainty-Aware ICRL (UAICRL) \cite{xu2024uncertaintyaware} tackles the ambiguity inherent in constraint inference by modeling both aleatoric and epistemic uncertainty, yielding risk-sensitive constraints via distributional Bellman updates and demonstrating robustness when demonstrations are limited or stochastic. Finally, CoCoRL \cite{lindner2024learning} bypasses reward labels and learns a convex safe set directly from demonstrations with unknown rewards, providing safety guarantees and transfer to new tasks. A key distinction from the ICRL family is that we do not infer an explicit constraint set or solve a constrained MDP. Instead, we regularize the value function directly: demonstrations define support, a discriminator gates in/out-of-distribution weighting, and a state-level upper bound limits over-optimism off-support while preserving SAC-style improvement on-demonstrations-support. This sidesteps the ambiguity and calibration issues inherent in constraint identification while still leveraging demonstrations to shape safe behavior.

\textbf{Offline RL with pessimistic value learning}. Offline RL addresses distribution shift by making critics conservative off-demo-nstrations-support. CQL \cite{kumar2020conservative} lowers Q-values on OOD actions to obtain lower-bound critics and policy-improvement guarantees. BRAC \cite{wu2020behavior} constrains the policy toward the behavior policy, while IQL \cite{kostrikov2021offline} uses expectiles to avoid querying OOD actions. Finally, model-based methods, such as MOPO \cite{NEURIPS2020_a322852c}, MOReL \cite{NEURIPS2020_f7efa4f8}, and COMBO \cite{yu2021combo}, inject pessimism through dynamics or value regularization. Aligned with this conservative principle, our method is pessimistic off-demonstrations-support but defines “support” via demonstrations together with the online replay buffer. Importantly, the pessimism operates at the state level: we bound Q-values for out-of-distribution states using a discriminator-weighted, demonstration-informed upper bound, so every action at those states is conservatively valued, while standard SAC updates drive improvement on in-distribution states. This contrasts with action-centric penalties such as CQL’s conservative regularizer and complements state-space pessimism in model-based approaches like MOReL/MOPO.


\section{Conclusions}

We introduced SafeQIL, a simple, model-free method for learning to respect constraints from demonstrations while continuing to improve online. The key idea is to make the critic pessimistic at out-of-distribution states by enforcing a local, demonstration-limited upper bound on $Q(s, \cdot)$ and by gating updates with a discriminator that estimates whether a state lies on the demonstration support. On in-distribution states, the policy still learns with a standard max-entropy objective (SAC), preserving the sample-efficiency and stability benefits of off-policy actor–critic training.

SafeQIL consistently reduces safety-violation costs compared to explicit constraint-inference baselines (ICRL/VICRL) in 4 Safety-Gymnasium tasks,  while maintaining competitive performance on navigation tasks. The method trades some reward for safety on interaction-heavy tasks, an expected outcome of our state-level pessimism design. These results support our central claim: shaping values to be conservative only where the data indicate low support yields strong safety without discarding the performance advantages of max-entropy RL.

Limitations include reliance on the coverage and quality of demonstrations, possible miscalibration of the discriminator in hard out-of-distribution regions, and the simplicity of the closest state retrieval for the anchor. These can lead to over- or under-conservatism in poorly covered parts of the state space.

Future work aims at (i) learning task-aware state embeddings for robust closest-demo retrieval, and (ii) exploring model-based rollouts for safer recovery planning.

\begin{acks}
This research work was supported by the Hellenic Foundation for Research and Innovation (HFRI) under the 5th Call for HFRI PhD Fellowships (Fellowship Number: 20769).
\end{acks}



\bibliographystyle{ACM-Reference-Format} 
\bibliography{sample}


\newpage
\onecolumn

\appendix

\section{Quantitative Trade-off Analysis}\label{app:quant_analysis}

In this section, we analytically describe how we select the best algorithm for each benchmark, and we present more table results. To do so, we introduce a quantitative trade-off analysis. This analysis measures how effectively each algorithm balances the drop in reward with the drop in cost relative to the unconstrained SAC baseline. We conduct this evaluation in two ways: a \textit{Standard} analysis based on mean performance, and a \textit{Robust} analysis that accounts for the variance (standard deviation) to penalize instability.

For a given algorithm $A$ and the unconstrained baseline $B$ (SAC), let $\mu_R, \sigma_R$ denote the mean and standard deviation of the reward, and $\mu_C, \sigma_C$ denote the mean and standard deviation of the cost. The \textit{Standard} metrics rely solely on the expected values. We define the percentage cost reduction ($\Delta C$) and reward reduction ($\Delta R$) as:
\begin{equation}
    \Delta C = \frac{\mu_C^B - \mu_C^A}{\mu_C^B} \times 100, \quad \Delta R = \frac{\mu_R^B - \mu_R^A}{\mu_R^B} \times 100
\end{equation}
The \textit{Trade-off Ratio} ($\rho$) is defined as the safety gained per unit of reward lost:
\begin{equation}
    \rho = \frac{\Delta C}{\Delta R}
\end{equation}
A higher $\rho$ indicates a more efficient safety mechanism. Algorithms with $\Delta C < 0$ (worse safety than baseline) are considered invalid.

To account for safety-critical constraints where worst-case performance matters, we define "pessimistic" bounds:
\begin{align}
    \tilde{C} &= \mu_C + \sigma_C \quad \text{(Upper Bound of Cost)} \\
    \tilde{R} &= \mu_R - \sigma_R \quad \text{(Lower Bound of Reward)}
\end{align}
The \textit{Robust} percentage drops are calculated using these worst-case estimates:
\begin{equation}
    \Delta \tilde{C} = \frac{\tilde{C}^B - \tilde{C}^A}{\tilde{C}^B} \times 100, \quad \Delta \tilde{R} = \frac{\tilde{R}^B - \tilde{R}^A}{\tilde{R}^B} \times 100
\end{equation}
This formulation penalizes algorithms with high variance, as a large $\sigma_C$ increases the worst-case cost estimate, thereby reducing the calculated safety improvement.

Table \ref{tab:standard_tradeoff} presents the \textit{Standard} trade-off analysis using mean values. Table \ref{tab:robust_tradeoff} presents the \textit{Robust} analysis, highlighting how variance impacts the reliability of safety guarantees. These results match the performance rankings discussed in Section \ref{sec:exp_res}. In SafetyPointGoal1-v0, SafeQIL achieves the highest trade-off ratio in both \textit{Standard} and \textit{Robust} results, confirming its superiority. In SafetyCarButton1-v0, while SAC-GAIL shows a marginally better ratio in the \textit{Standard} analysis ($0.68$ vs. $0.64$), the \textit{Robust} analysis reveals the necessity of accounting for variance. SafeQIL maintains a significantly tighter worst-case safety bound (Pessimistic Cost of $143.07$ vs. $179.69$ for SAC-GAIL) and achieves a better robust ratio ($0.39$ vs. $0.34$), justifying its selection as the most reliable algorithm for this benchmark. For SafetyPointCircle2-v0 and SafetyCarPush2-v0, where VICRL and SAC-GAIL achieve the highest ratios respectively, SafeQIL consistently ranks as the second-best performing algorithm at least in \textit{Robust} results. Crucially, while the baselines exhibit high variance in performance, that is, VICRL fails in SafetyCarPush2-v0 (Ratio $0.04$) and SAC-GAIL is unsafe in SafetyPointGoal1-v0, SafeQIL is the only method that maintains high efficacy and safety rankings across all benchmarks. This demonstrates that SafeQIL offers the most generalizable and robust solution, avoiding the catastrophic failures observed in other state-of-the-art methods.

\newpage

\begin{table}[H]
\caption{Standard Trade-off Analysis (Mean Values Only). The Ratio $\rho$ indicates \% Cost Drop per 1\% Reward Drop.}
\label{tab:standard_tradeoff}
\centering
\resizebox{\textwidth}{!}{
\begin{tabular}{llccccl}
\toprule
\textbf{Benchmark} & \textbf{Algorithm} & \textbf{Mean Reward} & \textbf{Mean Cost} & \textbf{Cost Drop ($\Delta C$)} & \textbf{Reward Drop ($\Delta R$)} & \textbf{Ratio ($\rho$)} \\
\midrule
\multirow{5}{*}{\shortstack{SafetyPointGoal1-v0\\(SAC: R 27.47, C 49.15)}} 
& \textbf{SafeQIL} & \textbf{5.27} & \textbf{34.22} & \textbf{30.4\%} & \textbf{80.8\%} & \textbf{0.38} \\
& ICRL$^2$ & 23.21 & 62.60 & -27.4\% & 15.5\% & N/A (Unsafe) \\
& ICRL$^7$ & -0.50 & 42.80 & 12.9\% & 101.8\% & 0.13 \\
& VICRL & 21.87 & 62.97 & -28.1\% & 20.4\% & N/A (Unsafe) \\
& SAC-GAIL & 7.17 & 44.80 & 8.9\% & 73.9\% & 0.12 \\
\midrule
\multirow{5}{*}{\shortstack{SafetyPointCircle2-v0\\(SAC: R 58.81, C 392.20)}} 
& SafeQIL & 27.06 & 29.28 & 92.5\% & 54.0\% & 1.71 \\
& ICRL$^2$ & 28.36 & 177.88 & 54.6\% & 51.8\% & 1.06 \\
& ICRL$^7$ & 8.21 & 34.07 & 91.3\% & 86.0\% & 1.06 \\
& \textbf{VICRL} & \textbf{26.29} & \textbf{5.49} & \textbf{98.6\%} & \textbf{55.3\%} & \textbf{1.78} \\
& SAC-GAIL & 11.89 & 33.95 & 91.3\% & 79.8\% & 1.14 \\
\midrule
\multirow{5}{*}{\shortstack{SafetyCarButton1-v0\\(SAC: R 19.49, C 299.27)}} 
& SafeQIL & -3.81 & 70.11 & 76.6\% & 119.5\% & 0.64 \\
& ICRL$^2$ & 0.31 & 222.01 & 25.8\% & 98.4\% & 0.26 \\
& ICRL$^7$ & -15.83 & 145.94 & 51.2\% & 181.2\% & 0.28 \\
& VICRL & -14.12 & 34.65 & 88.4\% & 172.4\% & 0.51 \\
& \textbf{SAC-GAIL} & \textbf{0.14} & \textbf{98.15} & \textbf{67.2\%} & \textbf{99.3\%} & \textbf{0.68} \\
\midrule
\multirow{5}{*}{\shortstack{SafetyCarPush2-v0\\(SAC: R 1.50, C 245.36)}} 
& SafeQIL & -0.62 & 57.20 & 76.7\% & 141.3\% & 0.54 \\
& ICRL$^2$ & 0.00 & 203.10 & 17.2\% & 100.0\% & 0.17 \\
& ICRL$^7$ & 1.15 & 178.81 & 27.1\% & 23.3\% & 1.16 \\
& VICRL & -9.05 & 55.80 & 77.3\% & 703.3\% & 0.11 \\
& \textbf{SAC-GAIL} & \textbf{0.76} & \textbf{87.97} & \textbf{64.1\%} & \textbf{49.3\%} & \textbf{1.30} \\
\bottomrule
\end{tabular}
}
\end{table}

\begin{table}[H]
\caption{Robust Trade-off Analysis (Mean $\pm$ Std). Pessimistic metrics penalize high variance in cost and reward.}
\label{tab:robust_tradeoff}
\centering
\resizebox{\textwidth}{!}{
\begin{tabular}{llccccl}
\toprule
\textbf{Benchmark} & \textbf{Algorithm} & \textbf{Pess. Reward ($\tilde{R}$)} & \textbf{Pess. Cost ($\tilde{C}$)} & \textbf{Robust $\Delta \tilde{C}$} & \textbf{Robust $\Delta \tilde{R}$} & \textbf{Ratio ($\rho$)} \\
\midrule
\multirow{5}{*}{\shortstack{SafetyPointGoal1-v0\\(SAC Pess: R 27.26, C 51.36)}} 
& \textbf{SafeQIL} & \textbf{3.42} & \textbf{36.93} & \textbf{28.1\%} & \textbf{87.5\%} & \textbf{0.32} \\
& ICRL$^2$ & 21.93 & 72.47 & -41.1\% & 19.6\% & N/A (Unsafe) \\
& ICRL$^7$ & -5.43 & 78.23 & -52.3\% & 119.9\% & N/A (Unsafe) \\
& VICRL & 20.67 & 73.78 & -43.6\% & 24.2\% & N/A (Unsafe) \\
& SAC-GAIL & 5.52 & 63.40 & -23.4\% & 79.7\% & N/A (Unsafe) \\
\midrule
\multirow{5}{*}{\shortstack{SafetyPointCircle2-v0\\(SAC Pess: R 58.11, C 396.58)}} 
& SafeQIL & 22.96 & 37.69 & 90.5\% & 60.5\% & 1.50 \\
& ICRL$^2$ & 10.99 & 184.58 & 53.5\% & 81.1\% & 0.66 \\
& ICRL$^7$ & -8.86 & 50.81 & 87.2\% & 115.2\% & 0.76 \\
& \textbf{VICRL} & \textbf{25.60} & \textbf{6.30} & \textbf{98.4\%} & \textbf{55.9\%} & \textbf{1.76} \\
& SAC-GAIL & 2.34 & 61.97 & 84.4\% & 96.0\% & 0.88 \\
\midrule
\multirow{5}{*}{\shortstack{SafetyCarButton1-v0\\(SAC Pess: R 15.99, C 319.88)}} 
& \textbf{SafeQIL} & \textbf{-6.86} & \textbf{143.07} & \textbf{55.3\%} & \textbf{142.9\%} & \textbf{0.39} \\
& ICRL$^2$ & -1.92 & 395.62 & -23.7\% & 112.0\% & N/A (Unsafe) \\
& ICRL$^7$ & -28.81 & 169.48 & 47.0\% & 280.2\% & 0.17 \\
& VICRL & -24.95 & 58.01 & 81.9\% & 256.0\% & 0.32 \\
& SAC-GAIL & -4.58 & 179.69 & 43.8\% & 128.6\% & 0.34 \\
\midrule
\multirow{5}{*}{\shortstack{SafetyCarPush2-v0\\(SAC Pess: R 1.49, C 273.88)}} 
& SafeQIL & -1.27 & 87.22 & 68.2\% & 185.2\% & 0.37 \\
& ICRL$^2$ & -1.73 & 240.34 & 12.2\% & 216.1\% & 0.06 \\
& ICRL$^7$ & 0.66 & 239.50 & 12.6\% & 55.7\% & 0.23 \\
& VICRL & -17.63 & 127.25 & 53.5\% & 1283.2\% & 0.04 \\
& \textbf{SAC-GAIL} & \textbf{-0.16} & \textbf{98.54} & \textbf{64.0\%} & \textbf{110.7\%} & \textbf{0.58} \\
\bottomrule
\end{tabular}
}
\end{table}

\newpage
\section{More learning curves}\label{app:more_curves}

\begin{figure}[H]
  \centering
  \includegraphics[width=1.0\textwidth]{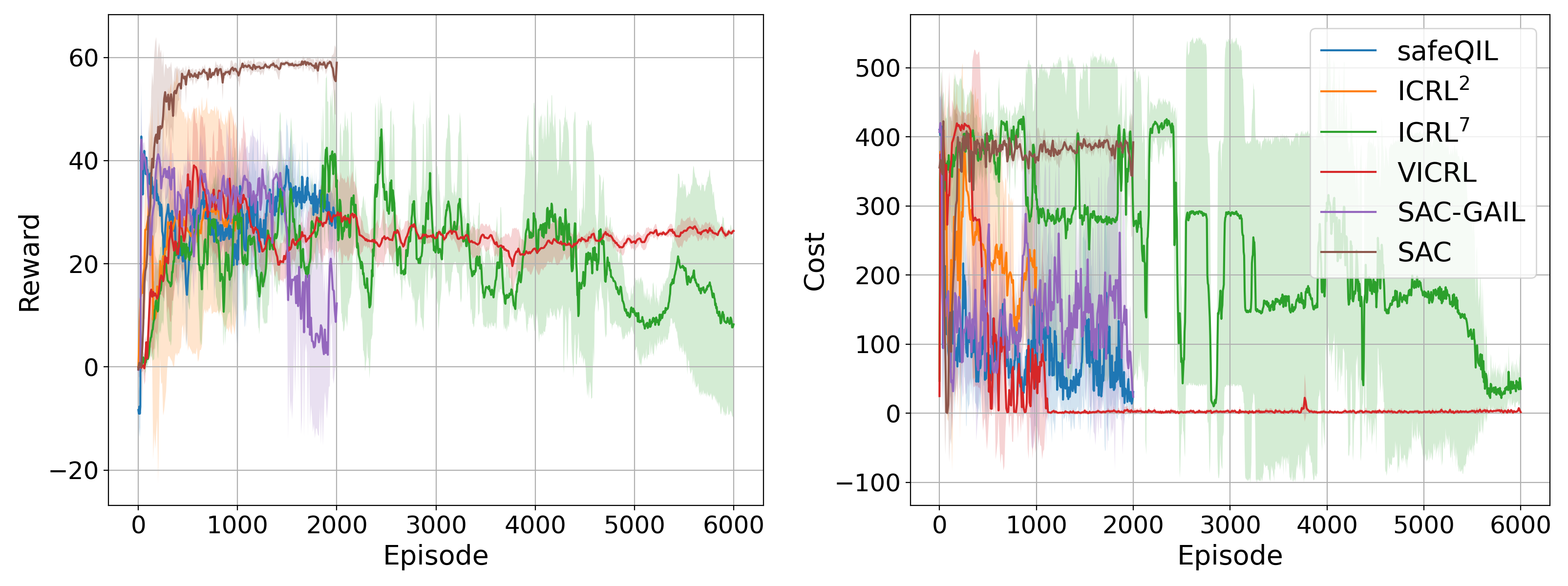}
  
  \caption{Learning curves for the SafetyPointCircle2-v0 benchmark.}
  \label{fig:pointcircle_exp}
\end{figure}

\begin{figure}[H]
  \centering
  \includegraphics[width=1.0\textwidth]{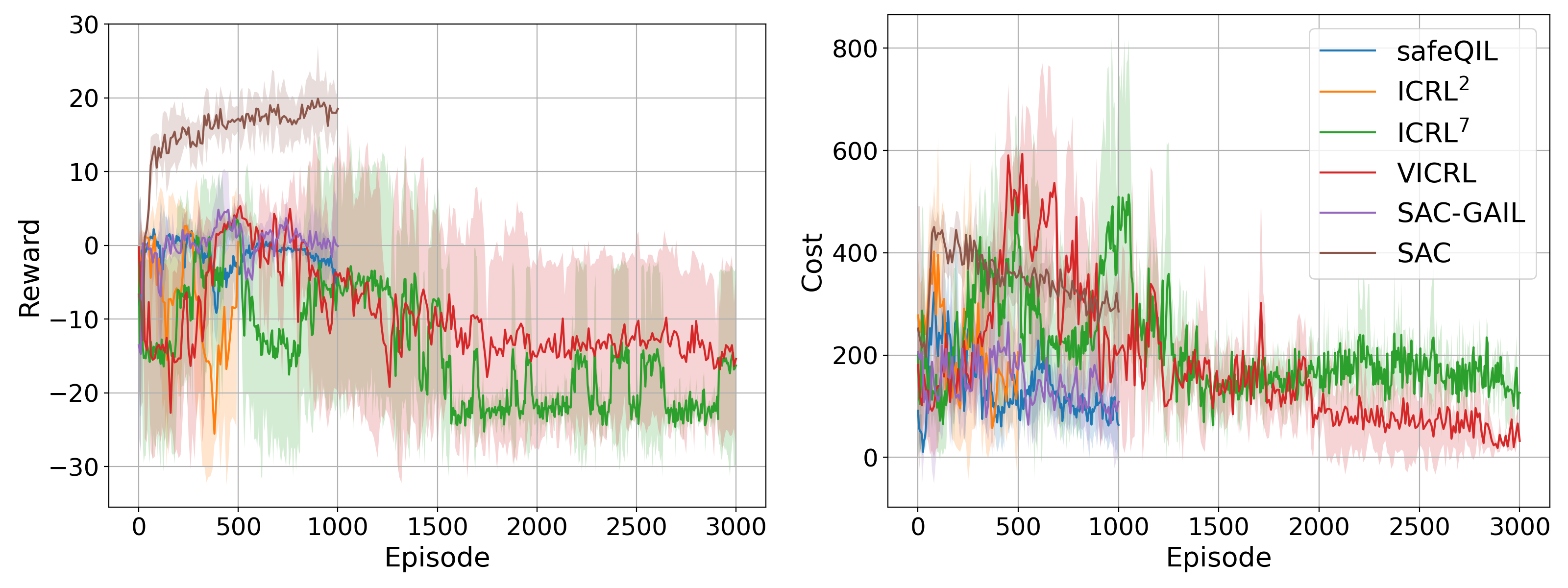}
  
  \caption{Learning curves for the SafetyCarButton1-v0 benchmark.}
  \label{fig:carbutton_exp}
\end{figure}

\newpage

\begin{figure}[H]
  \centering
  \includegraphics[width=1.0\textwidth]{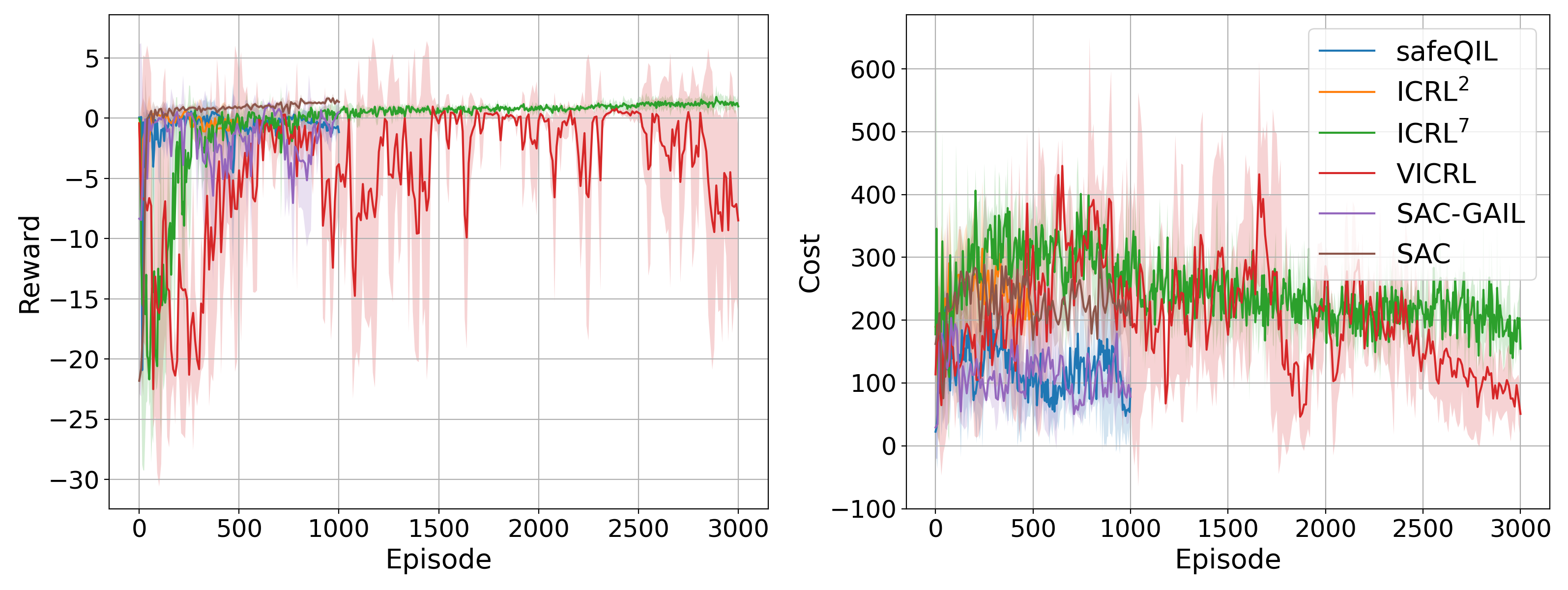}
  
  \caption{Learning curves for the SafetyCarPush2-v0 benchmark.}
  \label{fig:carpush_exp}
\end{figure}

\newpage

\section{Extended Ablation Study on SafeQIL Components}
\label{app:ablation_full}

In this section, we complete the ablation study initiated in the main text (Section \ref{sec:abl_study}) by evaluating the remaining components of SafeQIL's objective function. For completeness, Table \ref{tab:ablation_full} presents the full results across all 6 variations, including those previously discussed. 

First, removing the $\max(\cdot)$ operator ("w/o max term") in Equation (\ref{eq:final_constr-term}) forces the agent's value estimates to be strictly equal to the demonstrated bound rather than just staying below it. This severely restricts the policy, leading to a catastrophic drop in reward ($-10.62$), as removing the max term essentially prevents the critic from distinguishing the relative safety of different rollout states. Crucially, the removal of the Out-Of-Distribution term ("w/o OOD term"), i.e., the term $L^D_j(\cdot)$ defined in Equation (\ref{eq:final_ood-term}), resulted in the highest cost among all ablations ($44.99 \pm 4.95$). Without this term, the agent lacks an explicit safety penalty for states not covered by demonstrations, confirming that penalizing OOD states is central for SafeQIL's safety guarantees. Similarly, the demonstration regularization term plays a vital role in guiding the exploration. Removing it ("w/o demonstrations term") yielded a policy with negligible cost ($8.41$) but negative reward ($-3.71$), indicating that without initial guidance to follow safe trajectories, the agent collapses into a risk-averse only policy.

Finally, excluding the standard SAC objective ("w/o SAC term") hindered reward maximization ($2.67 \pm 1.78$). While the agent remained safe ($15.42 \pm 13.81$), it failed to exploit the safe regions effectively. In conclusion, the full study confirms that each component is necessary: OOD and demonstration terms help the agent identify safe regions, the SAC term drives performance, and the constraint mechanisms stabilize the learning process by providing robust upper bounds.

\begin{table}[H]
\caption{Complete ablation study evaluation results for SafeQIL.}
\label{tab:ablation_full}
\centering

\begin{tabular}{lcc}
\toprule
Algorithm & Reward~$\uparrow$ & Cost~$\downarrow$ \\

\midrule
\textbf{Original} & \textbf{5.27 $\pm$ 1.85} & \textbf{34.22 $\pm$ 2.71} \\
w/o cosine similarity & 3.74 $\pm$ 2.52 & 30.25 $\pm$ 19.10 \\
w/o max term & -10.62 $\pm$ 7.81 & 15.47 $\pm$ 8.67 \\
w/o constraint term & 4.34 $\pm$ 4.41 & 29.02 $\pm$ 14.03 \\
w/o OOD term & 1.07 $\pm$ 0.91 & 44.99 $\pm$ 4.95 \\
w/o demonstrations term & -3.71 $\pm$ 1.00 & 8.41 $\pm$ 3.76 \\
w/o SAC term & 2.67 $\pm$ 1.78 & 15.42 $\pm$ 13.81 \\

\bottomrule
\end{tabular}
\end{table}

\newpage
\section{Dataset Size Sensitivity Analysis}
\label{app:dataset_sensitivity}

In imitation learning and inverse reinforcement learning, increasing the volume of expert demonstration data typically correlates with improved policy stability and task performance. However, our ablation study across varying dataset sizes (1x, 2x, 4x, and 8x) reveals a counter-intuitive pattern. The performance of nearly all evaluated algorithms, measured by reward retention, worst-case cost bounds, and overall trade-off efficiency, generally degrades as the dataset size increases. This trend is observable in both the navigation task (SafetyPointGoal1-v0, Tables \ref{tab:abl-pointgooal-extend-data}, \ref{tab:std_tradeoff_ablation_pointgoal}, and \ref{tab:rob_tradeoff_ablation_pointgoal}) and the manipulation task (SafetyCarPush2-v0, Tables \ref{tab:abl-carpush-extend-data}, \ref{tab:std_tradeoff_ablation_carpush}, and \ref{tab:rob_tradeoff_ablation_carpush}), affecting direct value-regularization methods (SafeQIL), constraint inference methods (ICRL, VICRL), and standard imitation learning (SAC-GAIL).

Specifically, in SafetyPointGoal1-v0, SafeQIL achieves its highest robust trade-off efficiency at the 1x dataset size. As the dataset expands to 2x and 4x, this efficiency steadily drops, and by the 8x dataset size, reward variance increases significantly enough to push the pessimistic cost slightly above the unconstrained baseline threshold. A similar degradation is visible in ICRL and VICRL, which frequently fall into negative robust efficiency, effectively becoming more dangerous than the baseline, at larger dataset scales. In the complex SafetyCarPush2-v0 environment, SafeQIL provides a tight worst-case cost bound at 1x data, but this bound progressively weakens as more data is introduced. SAC-GAIL also experiences a severe efficiency drop at the 4x dataset size before uniquely recovering at the 8x scale.

We hypothesize that this counter-intuitive degradation is primarily driven by the stochasticity and multi-modal nature of human demonstrations. The datasets for these experiments were collected incrementally, for instance, the 2x dataset does not represent a single continuous session, but rather consists of the original 40 trajectories combined with an additional 40 trajectories collected at a later time. Over multiple days and collection sessions, a human demonstrator's strategy naturally drifts. The demonstrator may solve the exact same obstacle configuration using different trajectories, resulting in varying actions being mapped to highly similar state representations. When algorithms are trained on the 1x dataset, they fit a relatively narrow, consistent mode of behavior. However, as the dataset grows, the algorithms are forced to reconcile these conflicting demonstrations, creating severe approximation instability for standard Markovian policies.

This multi-modal behavior disproportionately impacts constraint inference methods. Because algorithms like ICRL and VICRL attempt to infer a single, unified constraint boundary, conflicting demonstration data causes the inferred constraints to become overly broad or entirely miscalibrated. This explains why these methods frequently exhibit catastrophic variance and task collapse at the 4x and 8x data sizes. The discriminator cannot reliably separate safe from unsafe regions when the expert data itself is inconsistent. SafeQIL mitigates this better in low-to-medium data regimes, but because it relies on a state-similarity metric to anchor its Q-values to demonstrated bounds, high stochasticity eventually causes the algorithm to retrieve conflicting Q-value bounds for similar states, leading to increased variance at the 8x scale. Conversely, SAC-GAIL struggles to provide reliable safety bounds in low-data regimes but shows recovery to some extend at 8x in both SafetyCarPush2-v0 and SafetyPointGoal1-v0, suggesting that adversarial imitation learning can eventually overcome demonstrator stochasticity by overwhelming the network with enough data to average out inconsistencies. However, this highlights a severe limitation in sample efficiency. Even when provided with the 8x dataset, SAC-GAIL remains vulnerable to worst-case safety violations and cannot match the reliable safety bounds achieved by SafeQIL using only the minimal 1x dataset.

To resolve the instability caused by human demonstrator drift, future iterations of these algorithms must address the inherent state aliasing present in multi-session datasets. A promising direction is the integration of Recurrent Neural Networks (RNNs) into the policy and critic architectures. By providing the network with a temporal history of the trajectory, the policy could distinguish between observationally similar states based on the agent's past momentum and context. This would allow the algorithm to correctly identify and adapt to the specific behavioral mode the human demonstrator was utilizing during a given session, stabilizing the learning process even as dataset sizes and demonstrator variance increase.

\begin{table}[H]
\caption{SafetyPointGoal1-v0: Comparative results of Dataset Size Ablation.}
\label{tab:abl-pointgooal-extend-data}
\centering
\begin{tabular}{lcc}
\toprule
\textbf{Algorithm} & \textbf{Reward}~$\uparrow$ & \textbf{Cost}~$\downarrow$ \\
\midrule
\multicolumn{3}{c}{\textit{1x Dataset}} \\
\textbf{SafeQIL} & \textbf{5.27 $\pm$ 1.85} & \textbf{34.22 $\pm$ 2.71} \\ 
ICRL$^2$ & 23.21 $\pm$ 1.28 & 62.60 $\pm$ 9.87 \\
ICRL$^7$ & -0.50 $\pm$ 4.93 & 42.80 $\pm$ 35.43 \\
VICRL & 21.87 $\pm$ 1.20 & 62.97 $\pm$ 10.81 \\
SAC-GAIL & 7.17 $\pm$ 1.65 & 44.80 $\pm$ 18.60 \\
\midrule
Human Demonstrator & 11.39 $\pm$ 1.55 & 0.00 $\pm$ 0.00 \\
\midrule
\multicolumn{3}{c}{\textit{2x Dataset}} \\
\textbf{SafeQIL} & \textbf{3.17 $\pm$ 0.95} & \textbf{29.50 $\pm$ 9.95} \\ 
ICRL$^2$ & 21.83 $\pm$ 2.31 & 66.17 $\pm$ 11.14 \\
ICRL$^7$ & -5.38 $\pm$ 5.28 & 22.16 $\pm$ 21.68 \\
VICRL & 0.42 $\pm$ 1.17 & 32.99 $\pm$ 13.74 \\
SAC-GAIL & 8.40 $\pm$ 5.25 & 47.00 $\pm$ 19.01 \\
\midrule
Human Demonstrator & 11.37 $\pm$ 1.72 & 0.00 $\pm$ 0.00 \\
\midrule
\multicolumn{3}{c}{\textit{4x Dataset}} \\
SafeQIL & 11.30 $\pm$ 1.95 & 34.14 $\pm$ 10.66 \\
ICRL$^2$ & 21.41 $\pm$ 2.09 & 60.38 $\pm$ 9.66 \\
\textbf{ICRL$^7$} & \textbf{-7.08 $\pm$ 2.85} & \textbf{6.64 $\pm$ 3.36} \\
VICRL & 21.10 $\pm$ 2.30 & 62.90 $\pm$ 14.29 \\
SAC-GAIL & 9.70 $\pm$ 1.88 & 46.39 $\pm$ 9.05 \\
\midrule
Human Demonstrator & 11.72 $\pm$ 1.72 & 0.00 $\pm$ 0.00 \\
\midrule
\multicolumn{3}{c}{\textit{8x Dataset}} \\
SafeQIL & 6.64 $\pm$ 6.76 & 43.91 $\pm$ 8.40 \\
ICRL$^2$ & 17.51 $\pm$ 3.63 & 66.55 $\pm$ 3.36 \\
ICRL$^7$ & 0.31 $\pm$ 2.19 & 46.05 $\pm$ 59.83 \\
VICRL & 12.83 $\pm$ 12.03 & 54.76 $\pm$ 29.31 \\
\textbf{SAC-GAIL} & \textbf{7.22 $\pm$ 1.98} & \textbf{40.47 $\pm$ 11.25} \\
\midrule
Human Demonstrator & 12.42 $\pm$ 1.96 & 0.00 $\pm$ 0.00 \\
\midrule
\multicolumn{3}{c}{\textit{Baselines}} \\
SAC & 27.47 $\pm$ 0.21 & 49.15 $\pm$ 2.21 \\
PPO$^2$ & 23.64 $\pm$ 1.64 & 60.99 $\pm$ 7.99 \\
PPO$^7$ & 26.70 $\pm$ 0.11 & 57.17 $\pm$ 1.68 \\
\bottomrule
\end{tabular}
\end{table}

\begin{table}[H]
\caption{SafetyCarPush2-v0: Comparative results of Dataset Size Ablation.}
\label{tab:abl-carpush-extend-data}
\centering
\begin{tabular}{lcc}
\toprule
\textbf{Algorithm} & \textbf{Reward}~$\uparrow$ & \textbf{Cost}~$\downarrow$ \\
\midrule
\multicolumn{3}{c}{\textit{1x Dataset}} \\
SafeQIL & -0.62 $\pm$ 0.65 & 57.20 $\pm$ 30.02 \\ 
ICRL$^2$ & 0.00 $\pm$ 1.73 & 203.10 $\pm$ 37.24 \\
ICRL$^7$ & 1.15 $\pm$ 0.49 & 178.81 $\pm$ 60.69 \\
VICRL & -9.05 $\pm$ 8.58 & 55.80 $\pm$ 71.45 \\
\textbf{SAC-GAIL} & \textbf{0.76 $\pm$ 0.92} & \textbf{87.97 $\pm$ 10.57} \\
\midrule
Human Demonstrator & 7.46 $\pm$ 1.77 & 0.00 $\pm$ 0.00 \\
\midrule
\multicolumn{3}{c}{\textit{2x Dataset}} \\
SafeQIL & -3.38 $\pm$ 2.97 & 58.10 $\pm$ 17.86 \\
ICRL$^2$ & 0.68 $\pm$ 0.35 & 206.67 $\pm$ 38.53 \\
ICRL$^7$ & 0.84 $\pm$ 0.60 & 203.40 $\pm$ 29.78 \\
VICRL & -0.55 $\pm$ 2.25 & 195.57 $\pm$ 54.00 \\
\textbf{SAC-GAIL} & \textbf{0.22 $\pm$ 0.47} & \textbf{90.95 $\pm$ 11.92} \\
\midrule
Human Demonstrator & 7.92 $\pm$ 2.01 & 0.00 $\pm$ 0.00 \\
\midrule
\multicolumn{3}{c}{\textit{4x Dataset}} \\
SafeQIL & -2.63 $\pm$ 3.28 & 100.13 $\pm$ 10.22 \\
ICRL$^2$ & -0.25 $\pm$ 0.96 & 231.99 $\pm$ 59.31 \\
ICRL$^7$ & -3.18 $\pm$ 7.15 & 228.45 $\pm$ 89.63 \\
VICRL & 0.36 $\pm$ 0.80 & 145.74 $\pm$ 89.95 \\
\textbf{SAC-GAIL} & \textbf{-0.62 $\pm$ 2.46} & \textbf{148.19 $\pm$ 24.23} \\
\midrule
Human Demonstrator & 8.59 $\pm$ 2.16 & 0.00 $\pm$ 0.00 \\
\midrule
\multicolumn{3}{c}{\textit{8x Dataset}} \\
SafeQIL & -1.67 $\pm$ 1.01 & 77.28 $\pm$ 44.47 \\
ICRL$^2$ & -2.82 $\pm$ 2.76 & 104.44 $\pm$ 60.95 \\
ICRL$^7$ & 1.05 $\pm$ 0.29 & 186.36 $\pm$ 33.88 \\
VICRL & -7.17 $\pm$ 7.15 & 97.10 $\pm$ 59.01 \\
\textbf{SAC-GAIL} & \textbf{1.07 $\pm$ 0.28} & \textbf{124.97 $\pm$ 20.58} \\
\midrule
Human Demonstrator & 8.96 $\pm$ 2.18 & 0.00 $\pm$ 0.00 \\
\midrule
\multicolumn{3}{c}{\textit{Baselines}} \\
SAC & 1.50 $\pm$ 0.01 & 245.36 $\pm$ 28.52 \\
PPO$^2$ & 0.44 $\pm$ 0.52 & 316.50 $\pm$ 52.84 \\
PPO$^7$ & 0.72 $\pm$ 0.64 & 238.63 $\pm$ 23.33 \\
\bottomrule
\end{tabular}
\end{table}

\begin{table}[H]
\caption{SafetyPointGoal1-v0: Standard Trade-off Analysis (Dataset Size Ablation). Calculated using Mean Values Only. The Ratio $\rho$ indicates \% Cost Drop per 1\% Reward Drop.}
\label{tab:std_tradeoff_ablation_pointgoal}
\centering
\resizebox{\textwidth}{!}{%
\begin{tabular}{llccccc}
\toprule
\textbf{Dataset Size} & \textbf{Algorithm} & \textbf{Mean Reward} & \textbf{Mean Cost} & \textbf{Cost Drop ($\Delta C$)} & \textbf{Reward Drop ($\Delta R$)} & \textbf{Ratio ($\rho$)} \\
\midrule
\multirow{5}{*}{\shortstack[l]{1x Dataset\\(SAC: R 27.47, C 49.15)}} 
& \textbf{SafeQIL} & \textbf{5.27} & \textbf{34.22} & \textbf{30.4\%} & \textbf{80.8\%} & \textbf{0.38} \\
& ICRL$^2$ & 23.21 & 62.60 & -27.4\% & 15.5\% & N/A (Unsafe) \\
& ICRL$^7$ & -0.50 & 42.80 & 12.9\% & 101.8\% & 0.13 \\
& VICRL & 21.87 & 62.97 & -28.1\% & 20.4\% & N/A (Unsafe) \\
& SAC-GAIL & 7.17 & 44.80 & 8.9\% & 73.9\% & 0.12 \\
\midrule
\multirow{5}{*}{\shortstack[l]{2x Dataset\\(SAC: R 27.47, C 49.15)}} 
& SafeQIL & 3.17 & 29.50 & 40.0\% & 88.5\% & 0.45 \\
& ICRL$^2$ & 21.83 & 66.17 & -34.6\% & 20.5\% & N/A (Unsafe) \\
& \textbf{ICRL$^7$} & \textbf{-5.38} & \textbf{22.16} & \textbf{54.9\%} & \textbf{119.6\%} & \textbf{0.46} \\
& VICRL & 0.42 & 32.99 & 32.9\% & 98.5\% & 0.33 \\
& SAC-GAIL & 8.40 & 47.00 & 4.4\% & 69.4\% & 0.06 \\
\midrule
\multirow{5}{*}{\shortstack[l]{4x Dataset\\(SAC: R 27.47, C 49.15)}} 
& SafeQIL & 11.30 & 34.14 & 30.5\% & 58.9\% & 0.52 \\
& ICRL$^2$ & 21.41 & 60.38 & -22.8\% & 22.1\% & N/A (Unsafe) \\
& \textbf{ICRL$^7$} & \textbf{-7.08} & \textbf{6.64} & \textbf{86.5\%} & \textbf{125.8\%} & \textbf{0.69} \\
& VICRL & 21.10 & 62.90 & -28.0\% & 23.2\% & N/A (Unsafe) \\
& SAC-GAIL & 9.70 & 46.39 & 5.6\% & 64.7\% & 0.09 \\
\midrule
\multirow{5}{*}{\shortstack[l]{8x Dataset\\(SAC: R 27.47, C 49.15)}} 
& SafeQIL & 6.64 & 43.91 & 10.7\% & 75.8\% & 0.14 \\
& ICRL$^2$ & 17.51 & 66.55 & -35.4\% & 36.3\% & N/A (Unsafe) \\
& ICRL$^7$ & 0.31 & 46.05 & 6.3\% & 98.9\% & 0.06 \\
& VICRL & 12.83 & 54.76 & -11.4\% & 53.3\% & N/A (Unsafe) \\
& \textbf{SAC-GAIL} & \textbf{7.22} & \textbf{40.47} & \textbf{17.7\%} & \textbf{73.7\%} & \textbf{0.24} \\
\bottomrule
\end{tabular}%
}
\end{table}

\begin{table}[H]
\caption{SafetyPointGoal1-v0: Robust Trade-off Analysis (Dataset Size Ablation). Calculated using worst-case bounds (Mean $\pm$ Std). Pessimistic metrics penalize high variance in cost and reward.}
\label{tab:rob_tradeoff_ablation_pointgoal}
\centering
\resizebox{\textwidth}{!}{%
\begin{tabular}{llccccc}
\toprule
\textbf{Dataset Size} & \textbf{Algorithm} & \textbf{Pess. Reward ($\tilde{R}$)} & \textbf{Pess. Cost ($\tilde{C}$)} & \textbf{Robust $\Delta\tilde{C}$} & \textbf{Robust $\Delta\tilde{R}$} & \textbf{Ratio ($\rho$)} \\
\midrule
\multirow{5}{*}{\shortstack[l]{1x Dataset\\(SAC Pess: R 27.26, C 51.36)}} 
& \textbf{SafeQIL} & \textbf{3.42} & \textbf{36.93} & \textbf{28.1\%} & \textbf{87.5\%} & \textbf{0.32} \\
& ICRL$^2$ & 21.93 & 72.47 & -41.1\% & 19.6\% & N/A (Unsafe) \\
& ICRL$^7$ & -5.43 & 78.23 & -52.3\% & 119.9\% & N/A (Unsafe) \\
& VICRL & 20.67 & 73.78 & -43.6\% & 24.2\% & N/A (Unsafe) \\
& SAC-GAIL & 5.52 & 63.40 & -23.4\% & 79.7\% & N/A (Unsafe) \\
\midrule
\multirow{5}{*}{\shortstack[l]{2x Dataset\\(SAC Pess: R 27.26, C 51.36)}} 
& \textbf{SafeQIL} & \textbf{2.22} & \textbf{39.45} & \textbf{23.2\%} & \textbf{91.9\%} & \textbf{0.25} \\
& ICRL$^2$ & 19.52 & 77.31 & -50.5\% & 28.4\% & N/A (Unsafe) \\
& ICRL$^7$ & -10.66 & 43.84 & 14.6\% & 139.1\% & 0.10 \\
& VICRL & -0.75 & 46.73 & 9.0\% & 102.8\% & 0.09 \\
& SAC-GAIL & 3.15 & 66.01 & -28.5\% & 88.4\% & N/A (Unsafe) \\
\midrule
\multirow{5}{*}{\shortstack[l]{4x Dataset\\(SAC Pess: R 27.26, C 51.36)}} 
& SafeQIL & 9.35 & 44.80 & 12.8\% & 65.7\% & 0.19 \\
& ICRL$^2$ & 19.32 & 70.04 & -36.4\% & 29.1\% & N/A (Unsafe) \\
& \textbf{ICRL$^7$} & \textbf{-9.93} & \textbf{10.00} & \textbf{80.5\%} & \textbf{136.4\%} & \textbf{0.59} \\
& VICRL & 18.80 & 77.19 & -50.3\% & 31.0\% & N/A (Unsafe) \\
& SAC-GAIL & 7.82 & 55.44 & -7.9\% & 71.3\% & N/A (Unsafe) \\
\midrule
\multirow{5}{*}{\shortstack[l]{8x Dataset\\(SAC Pess: R 27.26, C 51.36)}} 
& SafeQIL & -0.12 & 52.31 & -1.8\% & 100.4\% & N/A (Unsafe) \\
& ICRL$^2$ & 13.88 & 69.91 & -36.1\% & 49.1\% & N/A (Unsafe) \\
& ICRL$^7$ & -1.88 & 105.88 & -106.2\% & 106.9\% & N/A (Unsafe) \\
& VICRL & 0.80 & 84.07 & -63.7\% & 97.1\% & N/A (Unsafe) \\
& \textbf{SAC-GAIL} & \textbf{5.24} & \textbf{51.72} & \textbf{-0.7\%} & \textbf{80.8\%} & \textbf{N/A (Unsafe)} \\
\bottomrule
\end{tabular}
}
\end{table}

\begin{table}[H]
\caption{SafetyCarPush2-v0: Standard Trade-off Analysis (Dataset Size Ablation). Calculated using Mean Values Only. The Ratio $\rho$ indicates \% Cost Drop per 1\% Reward Drop.}
\label{tab:std_tradeoff_ablation_carpush}
\centering
\resizebox{\textwidth}{!}{%
\begin{tabular}{llccccc}
\toprule
\textbf{Dataset Size} & \textbf{Algorithm} & \textbf{Mean Reward} & \textbf{Mean Cost} & \textbf{Cost Drop ($\Delta C$)} & \textbf{Reward Drop ($\Delta R$)} & \textbf{Ratio ($\rho$)} \\
\midrule
\multirow{5}{*}{\shortstack[l]{1x Dataset\\(SAC: R 1.50, C 245.36)}} 
& SafeQIL & -0.62 & 57.20 & 76.7\% & 141.3\% & 0.54 \\
& ICRL$^2$ & 0.00 & 203.10 & 17.2\% & 100.0\% & 0.17 \\
& ICRL$^7$ & 1.15 & 178.81 & 27.1\% & 23.3\% & 1.16 \\
& VICRL & -9.05 & 55.80 & 77.3\% & 703.3\% & 0.11 \\
& \textbf{SAC-GAIL} & \textbf{0.76} & \textbf{87.97} & \textbf{64.1\%} & \textbf{49.3\%} & \textbf{1.30} \\
\midrule
\multirow{5}{*}{\shortstack[l]{2x Dataset\\(SAC: R 1.50, C 245.36)}} 
& SafeQIL & -3.38 & 58.10 & 76.3\% & 325.3\% & 0.23 \\
& ICRL$^2$ & 0.68 & 206.67 & 15.8\% & 54.7\% & 0.29 \\
& ICRL$^7$ & 0.84 & 203.40 & 17.1\% & 44.0\% & 0.39 \\
& VICRL & -0.55 & 195.57 & 20.3\% & 136.7\% & 0.15 \\
& \textbf{SAC-GAIL} & \textbf{0.22} & \textbf{90.95} & \textbf{62.9\%} & \textbf{85.3\%} & \textbf{0.74} \\
\midrule
\multirow{5}{*}{\shortstack[l]{4x Dataset\\(SAC: R 1.50, C 245.36)}} 
& SafeQIL & -2.63 & 100.13 & 59.2\% & 275.3\% & 0.21 \\
& ICRL$^2$ & -0.25 & 231.99 & 5.4\% & 116.7\% & 0.05 \\
& ICRL$^7$ & -3.18 & 228.45 & 6.9\% & 312.0\% & 0.02 \\
& \textbf{VICRL} & \textbf{0.36} & \textbf{145.74} & \textbf{40.6\%} & \textbf{76.0\%} & \textbf{0.53} \\
& SAC-GAIL & -0.62 & 148.19 & 39.6\% & 141.3\% & 0.28 \\
\midrule
\multirow{5}{*}{\shortstack[l]{8x Dataset\\(SAC: R 1.50, C 245.36)}} 
& SafeQIL & -1.67 & 77.28 & 68.5\% & 211.3\% & 0.32 \\
& ICRL$^2$ & -2.82 & 104.44 & 57.4\% & 288.0\% & 0.20 \\
& ICRL$^7$ & 1.05 & 186.36 & 24.0\% & 30.0\% & 0.80 \\
& VICRL & -7.17 & 97.10 & 60.4\% & 578.0\% & 0.10 \\
& \textbf{SAC-GAIL} & \textbf{1.07} & \textbf{124.97} & \textbf{49.1\%} & \textbf{28.7\%} & \textbf{1.71} \\
\bottomrule
\end{tabular}
}
\end{table}

\begin{table}[H]
\caption{SafetyCarPush2-v0: Robust Trade-off Analysis (Dataset Size Ablation). Calculated using worst-case bounds (Mean $\pm$ Std). Pessimistic metrics penalize high variance in cost and reward.}
\label{tab:rob_tradeoff_ablation_carpush}
\centering
\resizebox{\textwidth}{!}{%
\begin{tabular}{llccccc}
\toprule
\textbf{Dataset Size} & \textbf{Algorithm} & \textbf{Pess. Reward ($\tilde{R}$)} & \textbf{Pess. Cost ($\tilde{C}$)} & \textbf{Robust $\Delta\tilde{C}$} & \textbf{Robust $\Delta\tilde{R}$} & \textbf{Ratio ($\rho$)} \\
\midrule
\multirow{5}{*}{\shortstack[l]{1x Dataset\\(SAC Pess: R 1.49, C 273.88)}} 
& SafeQIL & -1.27 & 87.22 & 68.2\% & 185.2\% & 0.37 \\
& ICRL$^2$ & -1.73 & 240.34 & 12.2\% & 216.1\% & 0.06 \\
& ICRL$^7$ & 0.66 & 239.50 & 12.6\% & 55.7\% & 0.23 \\
& VICRL & -17.63 & 127.25 & 53.5\% & 1283.2\% & 0.04 \\
& \textbf{SAC-GAIL} & \textbf{-0.16} & \textbf{98.54} & \textbf{64.0\%} & \textbf{110.7\%} & \textbf{0.58} \\
\midrule
\multirow{5}{*}{\shortstack[l]{2x Dataset\\(SAC Pess: R 1.49, C 273.88)}} 
& SafeQIL & -6.35 & 75.96 & 72.3\% & 526.2\% & 0.14 \\
& ICRL$^2$ & 0.33 & 245.20 & 10.5\% & 77.9\% & 0.13 \\
& ICRL$^7$ & 0.24 & 233.18 & 14.9\% & 83.9\% & 0.18 \\
& VICRL & -2.80 & 249.57 & 8.9\% & 287.9\% & 0.03 \\
& \textbf{SAC-GAIL} & \textbf{-0.25} & \textbf{102.87} & \textbf{62.4\%} & \textbf{116.8\%} & \textbf{0.53} \\
\midrule
\multirow{5}{*}{\shortstack[l]{4x Dataset\\(SAC Pess: R 1.49, C 273.88)}} 
& \textbf{SafeQIL} & -5.91 & \textbf{110.35} & \textbf{59.7\%} & 496.6\% & \textbf{0.12} \\
& ICRL$^2$ & -1.21 & 291.30 & -6.4\% & 181.2\% & N/A (Unsafe) \\
& ICRL$^7$ & -10.33 & 318.08 & -16.1\% & 793.3\% & N/A (Unsafe) \\
& VICRL & -0.44 & 235.69 & 13.9\% & 129.5\% & 0.11 \\
& \textbf{SAC-GAIL} & \textbf{-3.08} & 172.42 & 37.0\% & \textbf{306.7\%} & \textbf{0.12} \\
\midrule
\multirow{5}{*}{\shortstack[l]{8x Dataset\\(SAC Pess: R 1.49, C 273.88)}} 
& SafeQIL & -2.68 & 121.75 & 55.5\% & 279.9\% & 0.20 \\
& ICRL$^2$ & -5.58 & 165.39 & 39.6\% & 474.5\% & 0.08 \\
& ICRL$^7$ & 0.76 & 220.24 & 19.6\% & 49.0\% & 0.40 \\
& VICRL & -14.32 & 156.11 & 43.0\% & 1061.1\% & 0.04 \\
& \textbf{SAC-GAIL} & \textbf{0.79} & \textbf{145.55} & \textbf{46.9\%} & \textbf{47.0\%} & \textbf{1.00} \\
\bottomrule
\end{tabular}
}
\end{table}

\newpage
\section{Hyperparameters}\label{app:hyperparameters}

In this section, we provide the hyperparameter settings of all algorithms used in our experiments. For SafeQIL, we used the hyperparameters presented in Table \ref{tab:safeqil-hyperparams} for all benchmarks and ablation studies, and we only tuned the Discriminator's Gradient Penalty Coefficient ($\lambda_{gp}$) for each case, since we found it to be the most critical hyperparameter. Specifically, we tried the set of values $\{0.05, 0.01, 0.005, 0.001, 0.0005, 0.0001\}$ for each case and we used $\lambda_{gp}=0.01$ for SafetyPointGoal1-v0 with x1 dataset size, $\lambda_{gp}=0.05$ for SafetyPointGoal1-v0 with x2 dataset size, $\lambda_{gp}=0.0005$ for SafetyPointGoal1-v0 with x4 dataset size, $\lambda_{gp}=0.0005$ for SafetyPointGoal1-v0 with x8 dataset size, $\lambda_{gp}=0.05$ for SafetyCarButton1-v0, $\lambda_{gp}=0.001$ for SafetyCarPush2-v0 with x1 dataset size, $\lambda_{gp}=0.01$ for SafetyCarPush2-v0 with x2 dataset size, $\lambda_{gp}=0.0005$ for SafetyCarPush2-v0 with x4 dataset size, $\lambda_{gp}=0.0001$ for SafetyCarPush2-v0 with x8 dataset size, $\lambda_{gp}=0.001$ for SafetyPointCircle2-v0, $\lambda_{gp}=0.05$ for "w/o cosine similarity" ablation, $\lambda_{gp}=0.01$ for "w/o max term" ablation, $\lambda_{gp}=0.005$ for "w/o constraint term" ablation,  $\lambda_{gp}=0.001$ for "w/o OOD term" ablation, $\lambda_{gp}=0.001$ for "w/o demonstrations term" ablation, and $\lambda_{gp}=0.01$ for "w/o SAC term" ablation.

\begin{table}[h]
\centering
\caption{Hyperparameter settings for SafeQIL.}
\label{tab:safeqil-hyperparams}
\begin{tabular}{lc}

\toprule

\textbf{Parameter} & \textbf{Value} \\
\midrule
\multicolumn{2}{l}{\textit{SAC (Actor / Critics)}} \\
\midrule

Actor/Critics Learning Rate & $3 \times 10^{-4}$ \\
Actor/Critics Optimizer & Adam \\
Discount Factor ($\gamma$) & $0.99$ \\
Batch Size & $256$ \\
Target Networks Smoothing Coefficient ($\eta$) & $0.005$ \\
Target Networks Update Interval & 1 \\
Entropy Coeff. ($\alpha$) Learning Rate & $3 \times 10^{-4}$ \\
Target Entropy ($\bar{\mathcal{H}}$) & $-\dim(A)$ \\
Initial Entropy Coefficient ($\alpha_0$) & 1.0 \\
Initial Log Std ($\log \sigma_{\text{init}}$) & -3.0 \\
Replay Buffer Size & $10^6$ \\
Total Environmnet Steps & $10^6$ \\
Warmup Environment Steps & $10^4$ \\
Gradient Steps per Update & $1$ \\
Environment Steps between Updates & $1$ \\
State-Dependent Exploration (SDE) & True \\
Use SDE at Warmup & False \\
Actor/Critics Hidden Layers & $32, 32$ \\
Actor/Critics Activation Function & ReLU \\

\midrule
\multicolumn{2}{l}{\textit{SafeQIL / Discriminator}} \\
\midrule

Discriminator Learning Rate & $3 \times 10^{-4}$ \\
Discriminator Optimizer & Adam \\
Discriminator Batch Size & $256$ \\
Discriminator Gradient Steps & $1$ \\
DAC Gradient Penalty Coefficient ($\lambda_{gp}$) & $0.005$ \\
Reward Function Type & $\log(\phi_{\omega}(s_B))$ \\
Discriminator Hidden Layers & $32, 32$ \\
Discriminator Activation Function & ReLU \\

\bottomrule
\end{tabular}
\end{table}

\newpage

For ICRL, we tried the 3 settings presented in Tables \ref{tab:icrl-hyperparams_1}, \ref{tab:icrl-hyperparams_2}, and \ref{tab:icrl-hyperparams_3}. At first, we tried them in the SafetyPointGoal1-v0 benchmark and we found that Setting 1 is the less promising. Then, we tried different values of Regularizer Weight ($\delta$) for Setting 2 and 3 in the same benchmark, using the set of values $\{0.8, 0.7, 0.6, 0.5, 0.4, 0.3, 0.2, 0.1\}$. We found that Setting 2 is the best one, thus we used it for all benchmarks, and we tuned only $\delta$ for each case using the same set of values. Specifically, we used $\delta=0.7$ for SafetyPointGoal1-v0 with x1, x2, and x4 dataset size, $\delta=0.8$ for SafetyPointGoal1-v0 with x8 dataset size, $\delta=0.1$ for SafetyCarButton1-v0, $\delta=0.8$ for SafetyCarPush2-v0 with x1 dataset size, $\delta=0.5$ for SafetyCarPush2-v0 with x2 dataset size, $\delta=0.6$ for SafetyCarPush2-v0 with x4 dataset size, $\delta=0.1$ for SafetyCarPush2-v0 with x8 dataset size, and $\delta=0.2$ for SafetyPointCircle2-v0. Furthermore, as mentioned in Section \ref{sec:exp_set}, we also tried the best VICRL setting (Setting 7) because these algorithms are very similar and they share the most hyperparameters. The only exception from the shared hyperparameters is the "Constraint Net $\zeta_{\theta}$ Output Activation Function" for which we keep ICRL's Sigmoid function instead of VICRL's Softplus since we consider it a central difference between these algorithms. Therefore, for the training of ICRL with VICRL Setting 7 we used $\delta=0.1$ for SafetyPointGoal1-v0 with x1 dataset size, $\delta=0.8$ for SafetyPointGoal1-v0 with x2 dataset size, $\delta=0.1$ for SafetyPointGoal1-v0 with x4 dataset size, $\delta=0.4$ for SafetyPointGoal1-v0 with x8 dataset size, $\delta=0.1$ for SafetyCarButton1-v0, $\delta=0.6$ for SafetyCarPush2-v0 with x1 dataset size, $\delta=0.7$ for SafetyCarPush2-v0 with x2 dataset size, $\delta=0.8$ for SafetyCarPush2-v0 with x4 and x8 dataset size, and $\delta=0.5$ for SafetyPointCircle2-v0. 

\begin{table}[H]
\centering
\caption{Hyperparameter setting 1 for ICRL.}
\label{tab:icrl-hyperparams_1}
\begin{tabular}{lc}
\toprule
\textbf{Parameter} & \textbf{Value} \\
\midrule
\multicolumn{2}{l}{\textit{General PPO (Forward Step)}} \\
\midrule
Policy/Critics Learning Rate & $3 \times 10^{-4}$ \\
Policy/Critics Optimizer & Adam \\
Reward GAE $\gamma$ & $0.99$ \\
Reward GAE $\lambda$ & $0.95$ \\
Reward Value Function Coefficient & $0.5$ \\
Max Gradient Norm & $0.5$ \\
Clip Range ($\epsilon$) & $0.2$ \\
Target KL ($\delta_{KL}$) & $0.01$ \\
Environment Steps per Update & $2000$ \\
Environment Steps per Iteration & $200000$ \\
Iterations ($N$) & $30$ \\
Batch Size & $64$ \\
Epochs per Policy/Critics Update & $20$ \\
Policy/Critics Hidden Layers & $64, 64$ \\
Policy/Critics Activation Function & Tanh \\
\midrule
\multicolumn{2}{l}{\textit{Constraint Learning (Backward Step)}} \\
\midrule
Constraint Network $\zeta_{\theta}$ Learning Rate & $0.01$ \\
Learning Rate Annealing Factor & $0.9$ \\
Constraint Network $\zeta_{\theta}$ Optimizer & Adam \\
Regularizer Weight ($\delta$) & $0.5$ \\
Max Forward KL ($\epsilon_F$) & $10$ \\
Max Backward KL ($\epsilon_B$) & $2.5$ \\
Cost GAE $\gamma$ & $0.99$ \\
Cost GAE $\lambda$ & $0.95$ \\
Cost Value Function Coefficient & $0.5$ \\
Rollouts per $\zeta_{\theta}$ Update ($M$) & $10$ \\
Epochs per $\zeta_{\theta}$ Update ($B$) & $10$ \\
Importance Sampling & True \\
Constraint Net $\zeta_{\theta}$ Hidden Layers & $20$ \\
Constraint Net $\zeta_{\theta}$ Activation Function & ReLU \\
Constraint Net $\zeta_{\theta}$ Output Activation Function & Sigmoid \\
\midrule
\multicolumn{2}{l}{\textit{Lagrangian $\lambda$}} \\
\midrule
Initial Value & $1.0$ \\
Learning Rate & $0.1$ \\
Optimizer & Adam \\
Activation Function & Softplus(Log($\lambda$)) \\
Cost Budget ($\alpha$) & $0.0$ \\
\bottomrule
\end{tabular}
\end{table}

\newpage

\begin{table}[H]
\centering
\caption{Hyperparameter setting 2 for ICRL.}
\label{tab:icrl-hyperparams_2}
\begin{tabular}{lc}
\toprule
\textbf{Parameter} & \textbf{Value} \\
\midrule
\multicolumn{2}{l}{\textit{General PPO (Forward Step)}} \\
\midrule
Policy/Critics Learning Rate & $3 \times 10^{-4}$ \\
Policy/Critics Optimizer & Adam \\
Reward GAE $\gamma$ & $0.99$ \\
Reward GAE $\lambda$ & $0.95$ \\
Reward Value Function Coefficient & $0.5$ \\
Max Gradient Norm & $0.5$ \\
Clip Range ($\epsilon$) & $0.2$ \\
Target KL ($\delta_{KL}$) & $0.01$ \\
Environment Steps per Update & $2000$ \\
Environment Steps per Iteration & $50000$ \\
Iterations ($N$) & $10$ \\
Batch Size & $64$ \\
Epochs per Policy/Critics Update & $20$ \\
Policy/Critics Hidden Layers & $64, 64$ \\
Policy/Critics Activation Function & Tanh \\
\midrule
\multicolumn{2}{l}{\textit{Constraint Learning (Backward Step)}} \\
\midrule
Constraint Network $\zeta_{\theta}$ Learning Rate & $0.01$ \\
Learning Rate Annealing Factor & $0.9$ \\
Constraint Network $\zeta_{\theta}$ Optimizer & Adam \\
Regularizer Weight ($\delta$) & $0.5$ \\
Max Forward KL ($\epsilon_F$) & $10$ \\
Max Backward KL ($\epsilon_B$) & $2.5$ \\
Cost GAE $\gamma$ & $0.99$ \\
Cost GAE $\lambda$ & $0.95$ \\
Cost Value Function Coefficient & $0.5$ \\
Rollouts per $\zeta_{\theta}$ Update ($M$) & $1$ \\
Epochs per $\zeta_{\theta}$ Update ($B$) & $10$ \\
Importance Sampling & True \\
Constraint Net $\zeta_{\theta}$ Hidden Layers & $20$ \\
Constraint Net $\zeta_{\theta}$ Activation Function & ReLU \\
Constraint Net $\zeta_{\theta}$ Output Activation Function & Sigmoid \\
\midrule
\multicolumn{2}{l}{\textit{Lagrangian $\lambda$}} \\
\midrule
Initial Value & $1.0$ \\
Learning Rate & $0.1$ \\
Optimizer & Adam \\
Activation Function & Softplus(Log($\lambda$)) \\
Cost Budget ($\alpha$) & $0.0$ \\
\bottomrule
\end{tabular}
\end{table}

\begin{table}[H]
\centering
\caption{Hyperparameter setting 3 for ICRL.}
\label{tab:icrl-hyperparams_3}
\begin{tabular}{lc}
\toprule
\textbf{Parameter} & \textbf{Value} \\
\midrule
\multicolumn{2}{l}{\textit{General PPO (Forward Step)}} \\
\midrule
Policy/Critics Learning Rate & $3 \times 10^{-5}$ \\
Policy/Critics Optimizer & Adam \\
Reward GAE $\gamma$ & $0.99$ \\
Reward GAE $\lambda$ & $0.9$ \\
Reward Value Function Coefficient & $0.5$ \\
Max Gradient Norm & $0.5$ \\
Clip Range ($\epsilon$) & $0.2$ \\
Target KL ($\delta_{KL}$) & $0.01$ \\
Environment Steps per Update & $2000$ \\
Environment Steps per Iteration & $200000$ \\
Iterations ($N$) & $20$ \\
Batch Size & $128$ \\
Epochs per Policy/Critics Update & $20$ \\
Policy/Critics Hidden Layers & $64, 64$ \\
Policy/Critics Activation Function & Tanh \\
\midrule
\multicolumn{2}{l}{\textit{Constraint Learning (Backward Step)}} \\
\midrule
Constraint Network $\zeta_{\theta}$ Learning Rate & $0.01$ \\
Learning Rate Annealing Factor & $0.9$ \\
Constraint Network $\zeta_{\theta}$ Optimizer & Adam \\
Regularizer Weight ($\delta$) & $0.6$ \\
Max Forward KL ($\epsilon_F$) & $10$ \\
Max Backward KL ($\epsilon_B$) & $2.5$ \\
Cost GAE $\gamma$ & $0.99$ \\
Cost GAE $\lambda$ & $0.95$ \\
Cost Value Function Coefficient & $0.5$ \\
Rollouts per $\zeta_{\theta}$ Update ($M$) & $45$ \\
Epochs per $\zeta_{\theta}$ Update ($B$) & $10$ \\
Importance Sampling & True \\
Constraint Net $\zeta_{\theta}$ Hidden Layers & $40, 40$ \\
Constraint Net $\zeta_{\theta}$ Activation Function & ReLU \\
Constraint Net $\zeta_{\theta}$ Output Activation Function & Sigmoid \\
\midrule
\multicolumn{2}{l}{\textit{Lagrangian $\lambda$}} \\
\midrule
Initial Value & $0.1$ \\
Learning Rate & $1.0$ \\
Optimizer & Adam \\
Activation Function & Softplus(Log($\lambda$)) \\
Cost Budget ($\alpha$) & $0.0$ \\
\bottomrule
\end{tabular}
\end{table}

\newpage

For VICRL, we tried in SafetyPointGoal1-v0 the 9 settings presented in Tables \ref{tab:Vicrl-hyperparams_1}-\ref{tab:Vicrl-hyperparams_9}. We found that Setting 7 yields the best results, thus we used it in all benchmarks, tuning only the hyperparameter Regularizer Weight ($\delta$) in each case using the set of values $\{0.8, 0.7, 0.6, 0.5, 0.4, 0.3, 0.2, 0.1\}$. Specifically, we used $\delta=0.6$ for SafetyPointGoal1-v0 with x1 dataset size, $\delta=0.1$ for SafetyPointGoal1-v0 with x2 dataset size, $\delta=0.7$ for SafetyPointGoal1-v0 with x4 dataset size, $\delta=0.2$ for SafetyPointGoal1-v0 with x8 dataset size, $\delta=0.1$ for SafetyCarButton1-v0, $\delta=0.1$ for SafetyCarPush2-v0 with x1 dataset size, $\delta=0.6$ for SafetyCarPush2-v0 with x2 dataset size, $\delta=0.4$ for SafetyCarPush2-v0 with x4 dataset size, $\delta=0.2$ for SafetyCarPush2-v0 with x8 dataset size, and $\delta=0.5$ for SafetyPointCircle2-v0.  

\begin{table}[H]
\centering
\caption{Hyperparameter setting 1 for VICRL.}
\label{tab:Vicrl-hyperparams_1}
\begin{tabular}{lc}
\toprule
\textbf{Parameter} & \textbf{Value} \\
\midrule
\multicolumn{2}{l}{\textit{General PPO (Forward Step)}} \\
\midrule
Policy/Critics Learning Rate & $3 \times 10^{-5}$ \\
Policy/Critics Optimizer & Adam \\
Reward GAE $\gamma$ & $0.99$ \\
Reward GAE $\lambda$ & $0.95$ \\
Reward Value Function Coefficient & $0.5$ \\
Max Gradient Norm & $0.5$ \\
Clip Range ($\epsilon$) & $0.2$ \\
Target KL ($\delta_{KL}$) & $0.01$ \\
Environment Steps per Update & $2000$ \\
Environment Steps per Iteration & $200000$ \\
Iterations ($N$) & $30$ \\
Batch Size & $64$ \\
Epochs per Policy/Critics Update & $10$ \\
Policy/Critics Hidden Layers & $64, 64$ \\
Policy/Critics Activation Function & Tanh \\
\midrule
\multicolumn{2}{l}{\textit{Constraint Learning (Backward Step)}} \\
\midrule
Constraint Network $\zeta_{\theta}$ Learning Rate & $0.005$ \\
Learning Rate Annealing Factor & $0.9$ \\
Constraint Network $\zeta_{\theta}$ Optimizer & Adam \\
Regularizer Weight ($\delta$) & $0.5$ \\
Max Forward KL ($\epsilon_F$) & $10$ \\
Max Backward KL ($\epsilon_B$) & $2.5$ \\
Cost GAE $\gamma$ & $0.99$ \\
Cost GAE $\lambda$ & $0.95$ \\
Cost Value Function Coefficient & $0.5$ \\
Rollouts per $\zeta_{\theta}$ Update ($M$) & $10$ \\
Epochs per $\zeta_{\theta}$ Update ($B$) & $10$ \\
Importance Sampling & True \\
Prior Dirichlet Distribution parameters $\boldsymbol{\alpha}_{\text{prior}}$ & [0.9, 0.1] \\
Constraint Net $\zeta_{\theta}$ Hidden Layers & $20$ \\
Constraint Net $\zeta_{\theta}$ Activation Function & ReLU \\
Constraint Net $\zeta_{\theta}$ Output Activation Function & Softplus \\
\midrule
\multicolumn{2}{l}{\textit{Lagrangian $\lambda$}} \\
\midrule
Initial Value & $1.0$ \\
Learning Rate & $0.01$ \\
Optimizer & Adam \\
Activation Function & Softplus(Log($\lambda$)) \\
Cost Budget ($\alpha$) & $0.0$ \\
\bottomrule
\end{tabular}
\end{table}

\newpage

\begin{table}[H]
\centering
\caption{Hyperparameter setting 2 for VICRL.}
\label{tab:Vicrl-hyperparams_2}
\begin{tabular}{lc}
\toprule
\textbf{Parameter} & \textbf{Value} \\
\midrule
\multicolumn{2}{l}{\textit{General PPO (Forward Step)}} \\
\midrule
Policy/Critics Learning Rate & $3 \times 10^{-5}$ \\
Policy/Critics Optimizer & Adam \\
Reward GAE $\gamma$ & $0.99$ \\
Reward GAE $\lambda$ & $0.95$ \\
Reward Value Function Coefficient & $0.5$ \\
Max Gradient Norm & $0.5$ \\
Clip Range ($\epsilon$) & $0.2$ \\
Target KL ($\delta_{KL}$) & $0.01$ \\
Environment Steps per Update & $2000$ \\
Environment Steps per Iteration & $200000$ \\
Iterations ($N$) & $30$ \\
Batch Size & $64$ \\
Epochs per Policy/Critics Update & $10$ \\
Policy/Critics Hidden Layers & $64, 64$ \\
Policy/Critics Activation Function & Tanh \\
\midrule
\multicolumn{2}{l}{\textit{Constraint Learning (Backward Step)}} \\
\midrule
Constraint Network $\zeta_{\theta}$ Learning Rate & $0.005$ \\
Learning Rate Annealing Factor & $0.9$ \\
Constraint Network $\zeta_{\theta}$ Optimizer & Adam \\
Regularizer Weight ($\delta$) & $0.5$ \\
Max Forward KL ($\epsilon_F$) & $10$ \\
Max Backward KL ($\epsilon_B$) & $2.5$ \\
Cost GAE $\gamma$ & $0.99$ \\
Cost GAE $\lambda$ & $0.95$ \\
Cost Value Function Coefficient & $0.5$ \\
Rollouts per $\zeta_{\theta}$ Update ($M$) & $10$ \\
Epochs per $\zeta_{\theta}$ Update ($B$) & $10$ \\
Importance Sampling & True \\
Prior Dirichlet Distribution parameters $\boldsymbol{\alpha}_{\text{prior}}$ & [9.0, 1.0] \\
Constraint Net $\zeta_{\theta}$ Hidden Layers & $20$ \\
Constraint Net $\zeta_{\theta}$ Activation Function & ReLU \\
Constraint Net $\zeta_{\theta}$ Output Activation Function & Softplus \\
\midrule
\multicolumn{2}{l}{\textit{Lagrangian $\lambda$}} \\
\midrule
Initial Value & $1.0$ \\
Learning Rate & $0.01$ \\
Optimizer & Adam \\
Activation Function & Softplus(Log($\lambda$)) \\
Cost Budget ($\alpha$) & $0.0$ \\
\bottomrule
\end{tabular}
\end{table}

\begin{table}[H]
\centering
\caption{Hyperparameter setting 3 for VICRL.}
\label{tab:Vicrl-hyperparams_3}
\begin{tabular}{lc}
\toprule
\textbf{Parameter} & \textbf{Value} \\
\midrule
\multicolumn{2}{l}{\textit{General PPO (Forward Step)}} \\
\midrule
Policy/Critics Learning Rate & $3 \times 10^{-5}$ \\
Policy/Critics Optimizer & Adam \\
Reward GAE $\gamma$ & $0.99$ \\
Reward GAE $\lambda$ & $0.9$ \\
Reward Value Function Coefficient & $0.5$ \\
Max Gradient Norm & $0.5$ \\
Clip Range ($\epsilon$) & $0.4$ \\
Target KL ($\delta_{KL}$) & $0.02$ \\
Environment Steps per Update & $2000$ \\
Environment Steps per Iteration & $200000$ \\
Iterations ($N$) & $100$ \\
Batch Size & $128$ \\
Epochs per Policy/Critics Update & $20$ \\
Policy/Critics Hidden Layers & $64, 64$ \\
Policy/Critics Activation Function & Tanh \\
\midrule
\multicolumn{2}{l}{\textit{Constraint Learning (Backward Step)}} \\
\midrule
Constraint Network $\zeta_{\theta}$ Learning Rate & $0.001$ \\
Learning Rate Annealing Factor & $0.9$ \\
Constraint Network $\zeta_{\theta}$ Optimizer & Adam \\
Regularizer Weight ($\delta$) & $0.6$ \\
Max Forward KL ($\epsilon_F$) & $10$ \\
Max Backward KL ($\epsilon_B$) & $2.5$ \\
Cost GAE $\gamma$ & $0.99$ \\
Cost GAE $\lambda$ & $0.9$ \\
Cost Value Function Coefficient & $0.5$ \\
Rollouts per $\zeta_{\theta}$ Update ($M$) & $50$ \\
Epochs per $\zeta_{\theta}$ Update ($B$) & $5$ \\
Importance Sampling & True \\
Prior Dirichlet Distribution parameters $\boldsymbol{\alpha}_{\text{prior}}$ & [0.009, 0.001] \\
Constraint Net $\zeta_{\theta}$ Hidden Layers & $64, 64$ \\
Constraint Net $\zeta_{\theta}$ Activation Function & ReLU \\
Constraint Net $\zeta_{\theta}$ Output Activation Function & Softplus \\
\midrule
\multicolumn{2}{l}{\textit{Lagrangian $\lambda$}} \\
\midrule
Initial Value & $0.1$ \\
Learning Rate & $0.05$ \\
Optimizer & Adam \\
Activation Function & Softplus(Log($\lambda$)) \\
Cost Budget ($\alpha$) & $0.0$ \\
\bottomrule
\end{tabular}
\end{table}

\begin{table}[H]
\centering
\caption{Hyperparameter setting 4 for VICRL.}
\label{tab:Vicrl-hyperparams_4}
\begin{tabular}{lc}
\toprule
\textbf{Parameter} & \textbf{Value} \\
\midrule
\multicolumn{2}{l}{\textit{General PPO (Forward Step)}} \\
\midrule
Policy/Critics Learning Rate & $3 \times 10^{-5}$ \\
Policy/Critics Optimizer & Adam \\
Reward GAE $\gamma$ & $0.99$ \\
Reward GAE $\lambda$ & $0.9$ \\
Reward Value Function Coefficient & $0.5$ \\
Max Gradient Norm & $0.5$ \\
Clip Range ($\epsilon$) & $0.4$ \\
Target KL ($\delta_{KL}$) & $0.01$ \\
Environment Steps per Update & $2000$ \\
Environment Steps per Iteration & $50000$ \\
Iterations ($N$) & $100$ \\
Batch Size & $128$ \\
Epochs per Policy/Critics Update & $20$ \\
Policy/Critics Hidden Layers & $64, 64$ \\
Policy/Critics Activation Function & Tanh \\
\midrule
\multicolumn{2}{l}{\textit{Constraint Learning (Backward Step)}} \\
\midrule
Constraint Network $\zeta_{\theta}$ Learning Rate & $0.001$ \\
Learning Rate Annealing Factor & $0.9$ \\
Constraint Network $\zeta_{\theta}$ Optimizer & Adam \\
Regularizer Weight ($\delta$) & $0.6$ \\
Max Forward KL ($\epsilon_F$) & $10$ \\
Max Backward KL ($\epsilon_B$) & $2.5$ \\
Cost GAE $\gamma$ & $0.99$ \\
Cost GAE $\lambda$ & $0.95$ \\
Cost Value Function Coefficient & $0.5$ \\
Rollouts per $\zeta_{\theta}$ Update ($M$) & $50$ \\
Epochs per $\zeta_{\theta}$ Update ($B$) & $5$ \\
Importance Sampling & True \\
Prior Dirichlet Distribution parameters $\boldsymbol{\alpha}_{\text{prior}}$ & [0.9, 0.1] \\
Constraint Net $\zeta_{\theta}$ Hidden Layers & $20$ \\
Constraint Net $\zeta_{\theta}$ Activation Function & ReLU \\
Constraint Net $\zeta_{\theta}$ Output Activation Function & Softplus \\
\midrule
\multicolumn{2}{l}{\textit{Lagrangian $\lambda$}} \\
\midrule
Initial Value & $5.0$ \\
Learning Rate & $0.01$ \\
Optimizer & Adam \\
Activation Function & Softplus(Log($\lambda$)) \\
Cost Budget ($\alpha$) & $0.0$ \\
\bottomrule
\end{tabular}
\end{table}

\begin{table}[H]
\centering
\caption{Hyperparameter setting 5 for VICRL.}
\label{tab:Vicrl-hyperparams_5}
\begin{tabular}{lc}
\toprule
\textbf{Parameter} & \textbf{Value} \\
\midrule
\multicolumn{2}{l}{\textit{General PPO (Forward Step)}} \\
\midrule
Policy/Critics Learning Rate & $3 \times 10^{-5}$ \\
Policy/Critics Optimizer & Adam \\
Reward GAE $\gamma$ & $0.99$ \\
Reward GAE $\lambda$ & $0.9$ \\
Reward Value Function Coefficient & $0.5$ \\
Max Gradient Norm & $0.5$ \\
Clip Range ($\epsilon$) & $0.4$ \\
Target KL ($\delta_{KL}$) & $0.02$ \\
Environment Steps per Update & $2000$ \\
Environment Steps per Iteration & $200000$ \\
Iterations ($N$) & $50$ \\
Batch Size & $128$ \\
Epochs per Policy/Critics Update & $20$ \\
Policy/Critics Hidden Layers & $64, 64$ \\
Policy/Critics Activation Function & Tanh \\
\midrule
\multicolumn{2}{l}{\textit{Constraint Learning (Backward Step)}} \\
\midrule
Constraint Network $\zeta_{\theta}$ Learning Rate & $0.005$ \\
Learning Rate Annealing Factor & $0.9$ \\
Constraint Network $\zeta_{\theta}$ Optimizer & Adam \\
Regularizer Weight ($\delta$) & $0.6$ \\
Max Forward KL ($\epsilon_F$) & $10$ \\
Max Backward KL ($\epsilon_B$) & $2.5$ \\
Cost GAE $\gamma$ & $0.99$ \\
Cost GAE $\lambda$ & $0.9$ \\
Cost Value Function Coefficient & $0.5$ \\
Rollouts per $\zeta_{\theta}$ Update ($M$) & $50$ \\
Epochs per $\zeta_{\theta}$ Update ($B$) & $5$ \\
Importance Sampling & True \\
Prior Dirichlet Distribution parameters $\boldsymbol{\alpha}_{\text{prior}}$ & [0.9, 0.1] \\
Constraint Net $\zeta_{\theta}$ Hidden Layers & $40, 40$ \\
Constraint Net $\zeta_{\theta}$ Activation Function & ReLU \\
Constraint Net $\zeta_{\theta}$ Output Activation Function & Softplus \\
\midrule
\multicolumn{2}{l}{\textit{Lagrangian $\lambda$}} \\
\midrule
Initial Value & $0.1$ \\
Learning Rate & $0.05$ \\
Optimizer & Adam \\
Activation Function & Softplus(Log($\lambda$)) \\
Cost Budget ($\alpha$) & $0.0$ \\
\bottomrule
\end{tabular}
\end{table}

\begin{table}[H]
\centering
\caption{Hyperparameter setting 6 for VICRL.}
\label{tab:Vicrl-hyperparams_6}
\begin{tabular}{lc}
\toprule
\textbf{Parameter} & \textbf{Value} \\
\midrule
\multicolumn{2}{l}{\textit{General PPO (Forward Step)}} \\
\midrule
Policy/Critics Learning Rate & $3 \times 10^{-5}$ \\
Policy/Critics Optimizer & Adam \\
Reward GAE $\gamma$ & $0.99$ \\
Reward GAE $\lambda$ & $0.9$ \\
Reward Value Function Coefficient & $0.5$ \\
Max Gradient Norm & $0.5$ \\
Clip Range ($\epsilon$) & $0.4$ \\
Target KL ($\delta_{KL}$) & $0.02$ \\
Environment Steps per Update & $2000$ \\
Environment Steps per Iteration & $20000$ \\
Iterations ($N$) & $500$ \\
Batch Size & $128$ \\
Epochs per Policy/Critics Update & $20$ \\
Policy/Critics Hidden Layers & $64, 64$ \\
Policy/Critics Activation Function & Tanh \\
\midrule
\multicolumn{2}{l}{\textit{Constraint Learning (Backward Step)}} \\
\midrule
Constraint Network $\zeta_{\theta}$ Learning Rate & $1 \times 10^{-4}$ \\
Learning Rate Annealing Factor & $0.9$ \\
Constraint Network $\zeta_{\theta}$ Optimizer & Adam \\
Regularizer Weight ($\delta$) & $0.6$ \\
Max Forward KL ($\epsilon_F$) & $10$ \\
Max Backward KL ($\epsilon_B$) & $2.5$ \\
Cost GAE $\gamma$ & $0.99$ \\
Cost GAE $\lambda$ & $0.9$ \\
Cost Value Function Coefficient & $0.5$ \\
Rollouts per $\zeta_{\theta}$ Update ($M$) & $50$ \\
Epochs per $\zeta_{\theta}$ Update ($B$) & $5$ \\
Importance Sampling & True \\
Prior Dirichlet Distribution parameters $\boldsymbol{\alpha}_{\text{prior}}$ & [0.09, 0.01] \\
Constraint Net $\zeta_{\theta}$ Hidden Layers & $20$ \\
Constraint Net $\zeta_{\theta}$ Activation Function & ReLU \\
Constraint Net $\zeta_{\theta}$ Output Activation Function & Softplus \\
\midrule
\multicolumn{2}{l}{\textit{Lagrangian $\lambda$}} \\
\midrule
Initial Value & $0.1$ \\
Learning Rate & $0.01$ \\
Optimizer & Adam \\
Activation Function & Softplus(Log($\lambda$)) \\
Cost Budget ($\alpha$) & $0.0$ \\
\bottomrule
\end{tabular}
\end{table}

\begin{table}[H]
\centering
\caption{Hyperparameter setting 7 for VICRL.}
\label{tab:Vicrl-hyperparams_7}
\begin{tabular}{lc}
\toprule
\textbf{Parameter} & \textbf{Value} \\
\midrule
\multicolumn{2}{l}{\textit{General PPO (Forward Step)}} \\
\midrule
Policy/Critics Learning Rate & $1 \times 10^{-4}$ \\
Policy/Critics Optimizer & Adam \\
Reward GAE $\gamma$ & $0.99$ \\
Reward GAE $\lambda$ & $0.95$ \\
Reward Value Function Coefficient & $0.5$ \\
Max Gradient Norm & $0.5$ \\
Clip Range ($\epsilon$) & $0.2$ \\
Target KL ($\delta_{KL}$) & $0.01$ \\
Environment Steps per Update & $2000$ \\
Environment Steps per Iteration & $6000$ \\
Iterations ($N$) & $500$ \\
Batch Size & $2000$ \\
Epochs per Policy/Critics Update & $10$ \\
Policy/Critics Hidden Layers & $64, 64$ \\
Policy/Critics Activation Function & Tanh \\
\midrule
\multicolumn{2}{l}{\textit{Constraint Learning (Backward Step)}} \\
\midrule
Constraint Network $\zeta_{\theta}$ Learning Rate & $5 \times 10^{-3}$ \\
Learning Rate Annealing Factor & $0.9$ \\
Constraint Network $\zeta_{\theta}$ Optimizer & Adam \\
Regularizer Weight ($\delta$) & $0.5$ \\
Max Forward KL ($\epsilon_F$) & $10$ \\
Max Backward KL ($\epsilon_B$) & $2.5$ \\
Cost GAE $\gamma$ & $0.99$ \\
Cost GAE $\lambda$ & $0.9$ \\
Cost Value Function Coefficient & $0.5$ \\
Rollouts per $\zeta_{\theta}$ Update ($M$) & $100$ \\
Epochs per $\zeta_{\theta}$ Update ($B$) & $10$ \\
Importance Sampling & True \\
Prior Dirichlet Distribution parameters $\boldsymbol{\alpha}_{\text{prior}}$ & [9.0, 1.0] \\
Constraint Net $\zeta_{\theta}$ Hidden Layers & $64, 64$ \\
Constraint Net $\zeta_{\theta}$ Batch Size & 500 \\
Constraint Net $\zeta_{\theta}$ Activation Function & ReLU \\
Constraint Net $\zeta_{\theta}$ Output Activation Function & Softplus \\
\midrule
\multicolumn{2}{l}{\textit{Lagrangian $\lambda$}} \\
\midrule
Initial Value & $1.0$ \\
Learning Rate & $0.01$ \\
Optimizer & Adam \\
Activation Function & Softplus(Log($\lambda$)) \\
Cost Budget ($\alpha$) & $0.0$ \\
\bottomrule
\end{tabular}
\end{table}

\begin{table}[H]
\centering
\caption{Hyperparameter setting 8 for VICRL.}
\label{tab:Vicrl-hyperparams_8}
\begin{tabular}{lc}
\toprule
\textbf{Parameter} & \textbf{Value} \\
\midrule
\multicolumn{2}{l}{\textit{General PPO (Forward Step)}} \\
\midrule
Policy/Critics Learning Rate & $1 \times 10^{-4}$ \\
Policy/Critics Optimizer & Adam \\
Reward GAE $\gamma$ & $0.99$ \\
Reward GAE $\lambda$ & $0.95$ \\
Reward Value Function Coefficient & $0.5$ \\
Max Gradient Norm & $0.5$ \\
Clip Range ($\epsilon$) & $0.2$ \\
Target KL ($\delta_{KL}$) & $0.01$ \\
Environment Steps per Update & $2000$ \\
Environment Steps per Iteration & $6000$ \\
Iterations ($N$) & $150$ \\
Batch Size & $2000$ \\
Epochs per Policy/Critics Update & $10$ \\
Policy/Critics Hidden Layers & $64, 64$ \\
Policy/Critics Activation Function & Tanh \\
\midrule
\multicolumn{2}{l}{\textit{Constraint Learning (Backward Step)}} \\
\midrule
Constraint Network $\zeta_{\theta}$ Learning Rate & $5 \times 10^{-3}$ \\
Learning Rate Annealing Factor & $0.5$ \\
Constraint Network $\zeta_{\theta}$ Optimizer & Adam \\
Regularizer Weight ($\delta$) & $0.5$ \\
Max Forward KL ($\epsilon_F$) & $10$ \\
Max Backward KL ($\epsilon_B$) & $2.5$ \\
Cost GAE $\gamma$ & $0.99$ \\
Cost GAE $\lambda$ & $0.95$ \\
Cost Value Function Coefficient & $0.5$ \\
Rollouts per $\zeta_{\theta}$ Update ($M$) & $100$ \\
Epochs per $\zeta_{\theta}$ Update ($B$) & $10$ \\
Importance Sampling & True \\
Prior Dirichlet Distribution parameters $\boldsymbol{\alpha}_{\text{prior}}$ & [0.9, 0.1] \\
Constraint Net $\zeta_{\theta}$ Hidden Layers & $20$ \\
Constraint Net $\zeta_{\theta}$ Batch Size & 500 \\
Constraint Net $\zeta_{\theta}$ Activation Function & ReLU \\
Constraint Net $\zeta_{\theta}$ Output Activation Function & Softplus \\
\midrule
\multicolumn{2}{l}{\textit{Lagrangian $\lambda$}} \\
\midrule
Initial Value & $1.0$ \\
Learning Rate & $0.01$ \\
Optimizer & Adam \\
Activation Function & Softplus(Log($\lambda$)) \\
Cost Budget ($\alpha$) & $0.0$ \\
\bottomrule
\end{tabular}
\end{table}

\begin{table}[H]
\centering
\caption{Hyperparameter setting 9 for VICRL.}
\label{tab:Vicrl-hyperparams_9}
\begin{tabular}{lc}
\toprule
\textbf{Parameter} & \textbf{Value} \\
\midrule
\multicolumn{2}{l}{\textit{General PPO (Forward Step)}} \\
\midrule
Policy/Critics Learning Rate & $1 \times 10^{-4}$ \\
Policy/Critics Optimizer & Adam \\
Reward GAE $\gamma$ & $0.99$ \\
Reward GAE $\lambda$ & $0.95$ \\
Reward Value Function Coefficient & $0.5$ \\
Max Gradient Norm & $0.5$ \\
Clip Range ($\epsilon$) & $0.2$ \\
Target KL ($\delta_{KL}$) & $0.01$ \\
Environment Steps per Update & $2000$ \\
Environment Steps per Iteration & $6000$ \\
Iterations ($N$) & $500$ \\
Batch Size & $2000$ \\
Epochs per Policy/Critics Update & $10$ \\
Policy/Critics Hidden Layers & $64, 64$ \\
Policy/Critics Activation Function & Tanh \\
\midrule
\multicolumn{2}{l}{\textit{Constraint Learning (Backward Step)}} \\
\midrule
Constraint Network $\zeta_{\theta}$ Learning Rate & $5 \times 10^{-4}$ \\
Learning Rate Annealing Factor & $0.5$ \\
Constraint Network $\zeta_{\theta}$ Optimizer & Adam \\
Regularizer Weight ($\delta$) & $0.5$ \\
Max Forward KL ($\epsilon_F$) & $10$ \\
Max Backward KL ($\epsilon_B$) & $2.5$ \\
Cost GAE $\gamma$ & $0.99$ \\
Cost GAE $\lambda$ & $0.95$ \\
Cost Value Function Coefficient & $0.5$ \\
Rollouts per $\zeta_{\theta}$ Update ($M$) & $100$ \\
Epochs per $\zeta_{\theta}$ Update ($B$) & $10$ \\
Importance Sampling & True \\
Prior Dirichlet Distribution parameters $\boldsymbol{\alpha}_{\text{prior}}$ & [0.9, 0.1] \\
Constraint Net $\zeta_{\theta}$ Hidden Layers & $20$ \\
Constraint Net $\zeta_{\theta}$ Batch Size & 1000 \\
Constraint Net $\zeta_{\theta}$ Activation Function & ReLU \\
Constraint Net $\zeta_{\theta}$ Output Activation Function & Softplus \\
\midrule
\multicolumn{2}{l}{\textit{Lagrangian $\lambda$}} \\
\midrule
Initial Value & $1.0$ \\
Learning Rate & $0.01$ \\
Optimizer & Adam \\
Activation Function & Softplus(Log($\lambda$)) \\
Cost Budget ($\alpha$) & $0.0$ \\
\bottomrule
\end{tabular}
\end{table}

\newpage

For SAC-GAIL, we used the setting presented in Table \ref{tab:sac-gail-hyperparams} and we only tuned the hyperparameter Gradient Penalty Coefficient ($\lambda_{gp}$) for each case, using the set of values $\{0.05, 0.01, 0.005, 0.001, 0.0005, 0.0001\}$. Specifically, we used $\lambda_{gp}=0.0001$ for SafetyPointGoal1-v0 with x1 dataset size, $\lambda_{gp}=0.001$ for SafetyPointGoal1-v0 with x2 dataset size, $\lambda_{gp}=0.05$ for SafetyPointGoal1-v0 with x4 dataset size, $\lambda_{gp}=0.0005$ for SafetyPointGoal1-v0 with x8 dataset size, $\lambda_{gp}=0.001$ for SafetyCarButton1-v0, $\lambda_{gp}=0.01$ for SafetyCarPush2-v0 with x1 dataset size, $\lambda_{gp}=0.05$ for SafetyCarPush2-v0 with x2 dataset size, $\lambda_{gp}=0.005$ for SafetyCarPush2-v0 with x4 dataset size, $\lambda_{gp}=0.01$ for SafetyCarPush2-v0 with x8 dataset size, and $\lambda_{gp}=0.01$ for SafetyPointCircle2-v0.

\begin{table}[H]
\centering
\caption{Hyperparameter settings for SAC-GAIL.}
\label{tab:sac-gail-hyperparams}
\begin{tabular}{lc}

\toprule

\textbf{Parameter} & \textbf{Value} \\
\midrule
\multicolumn{2}{l}{\textit{SAC (Actor / Critics)}} \\
\midrule

Actor/Critics Learning Rate & $3 \times 10^{-4}$ \\
Actor/Critics Optimizer & Adam \\
Discount Factor ($\gamma$) & $0.99$ \\
Batch Size & $256$ \\
Target Networks Smoothing Coefficient ($\eta$) & $0.005$ \\
Target Networks Update Interval & 1 \\
Entropy Coeff. ($\alpha$) Learning Rate & $3 \times 10^{-4}$ \\
Target Entropy ($\bar{\mathcal{H}}$) & $-\dim(A)$ \\
Initial Entropy Coefficient ($\alpha_0$) & 1.0 \\
Initial Log Std ($\log \sigma_{\text{init}}$) & -3.0 \\
Replay Buffer Size & $10^6$ \\
Total Environmnet Steps & $10^6$ \\
Warmup Environment Steps & $10^4$ \\
Gradient Steps per Update & $1$ \\
Environment Steps between Updates & $1$ \\
State-Dependent Exploration (SDE) & True \\
Use SDE at Warmup & False \\
Actor/Critics Hidden Layers & $32, 32$ \\
Actor/Critics Activation Function & ReLU \\

\midrule
\multicolumn{2}{l}{\textit{GAIL / Discriminator}} \\
\midrule

Discriminator Learning Rate & $3 \times 10^{-4}$ \\
Discriminator Optimizer & Adam \\
Discriminator Batch Size & $256$ \\
Discriminator Gradient Steps & $1$ \\
DAC Gradient Penalty Coefficient ($\lambda_{gp}$) & $0.005$ \\
Reward Function Type & $\log(\phi_{\omega}(s_B, a_B))$ \\
Discriminator Hidden Layers & $32, 32$ \\
Discriminator Activation Function & ReLU \\

\bottomrule
\end{tabular}
\end{table}

\end{document}